\RequirePackage{fix-cm}
\documentclass[twocolumn]{svjour3}     
\smartqed 
\usepackage{graphicx}
\usepackage[caption=false,font=footnotesize,labelfont=sf,textfont=sf]{subfig}
\usepackage{amsmath}
\usepackage{amssymb}
\usepackage{algorithm}
\usepackage{algorithmicx}
\usepackage[square,numbers]{natbib}
\usepackage{url}
\usepackage{tcolorbox}

\newcommand{\real}{\ensuremath{\mathbb{R}}}

\newtheorem{defn}{Definition}

\begin{document}

\title{A Quotient Space Formulation for Generative Statistical Analysis of Graphical Data
}


\author{Xiaoyang Guo     \and
    		 Anuj~Srivastava   \and
    		 Sudeep~Sarkar	
}


\institute{X. Guo and A. Srivastava \at
				Department of Statistics, Florida State University, Tallahassee, FL 32306, USA.\\       				\email{\{xiaoyang.guo,anuj\}@stat.fsu.edu}      
      \and
      S. Sarkar \at
      Department of Computer Science and Engineering, University of South Florida, Tampa, FL 33620, USA.\\
      \email{sarkar@usf.edu}
}

\date{Received: date / Accepted: date}

\maketitle

\begin{abstract}
Complex analyses involving multiple, dependent random quantities
often lead to graphical models -- a set of nodes 
denoting variables of interest, and corresponding edges denoting statistical interactions between 
nodes. To develop statistical analyses for graphical data, especially towards generative modeling, 
one needs mathematical 
representations and metrics
for matching and comparing graphs, 
and subsequent tools, such as geodesics, means, and covariances. This paper utilizes a quotient structure to develop 
efficient algorithms for computing these quantities, leading to useful statistical 
tools, including principal component 
analysis, statistical testing, and modeling.  
We demonstrate the efficacy of this framework using datasets taken from several
problem areas, including letters, biochemical 
structures, and social networks.
\end{abstract}

\section{Introduction}
\label{intro}
\begin{figure} 
\centering
\subfloat[Social Network]
{
\includegraphics[width=0.5\linewidth]{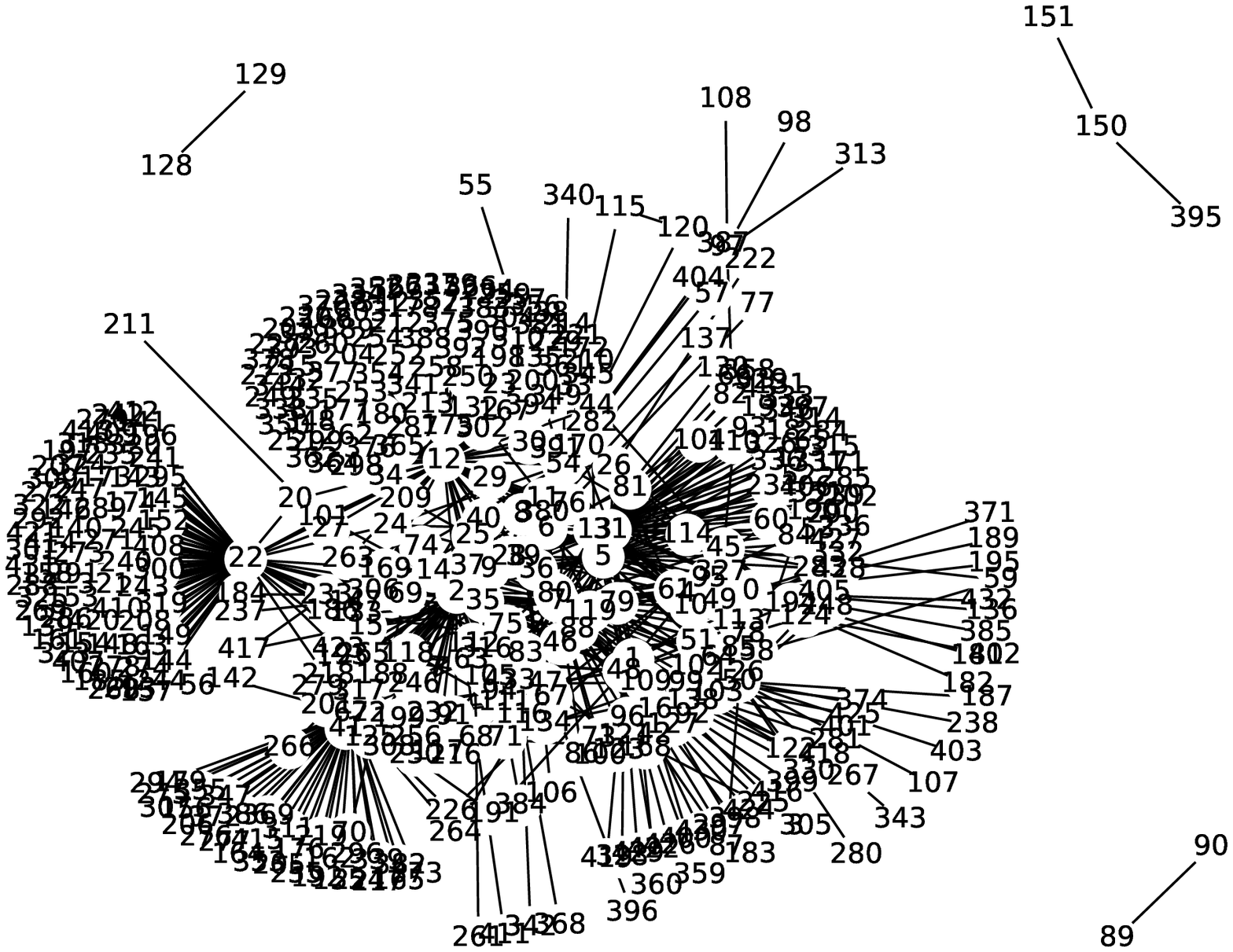}   
}
\subfloat[Molecule]
{
\includegraphics[width=0.5\linewidth]{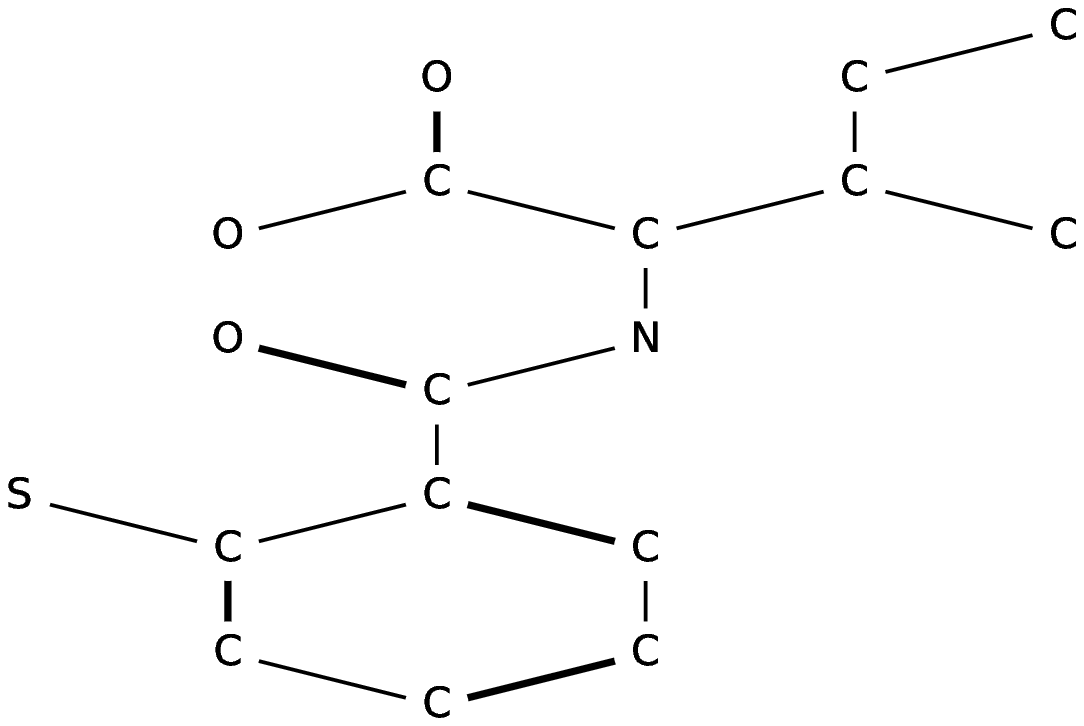}   
}\\
\subfloat[Video Interpretation]
{
\includegraphics[width=0.5\linewidth]{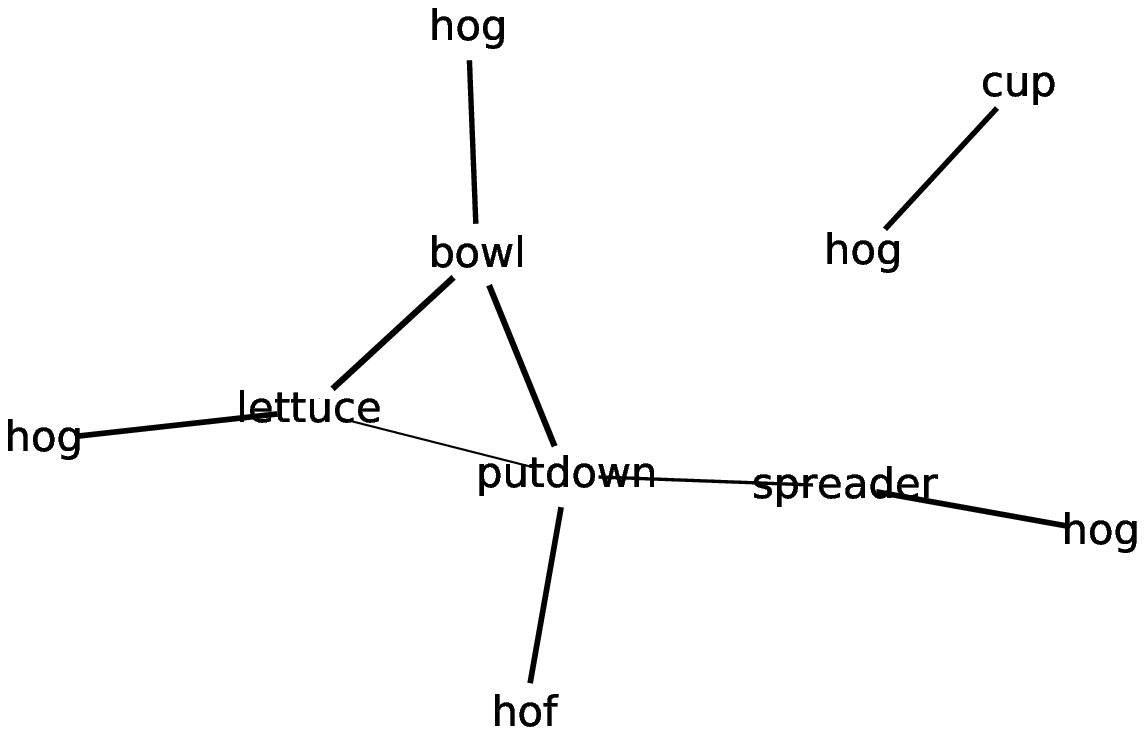}   
}
\subfloat[Brain Artery]
{
\includegraphics[width=0.5\linewidth]{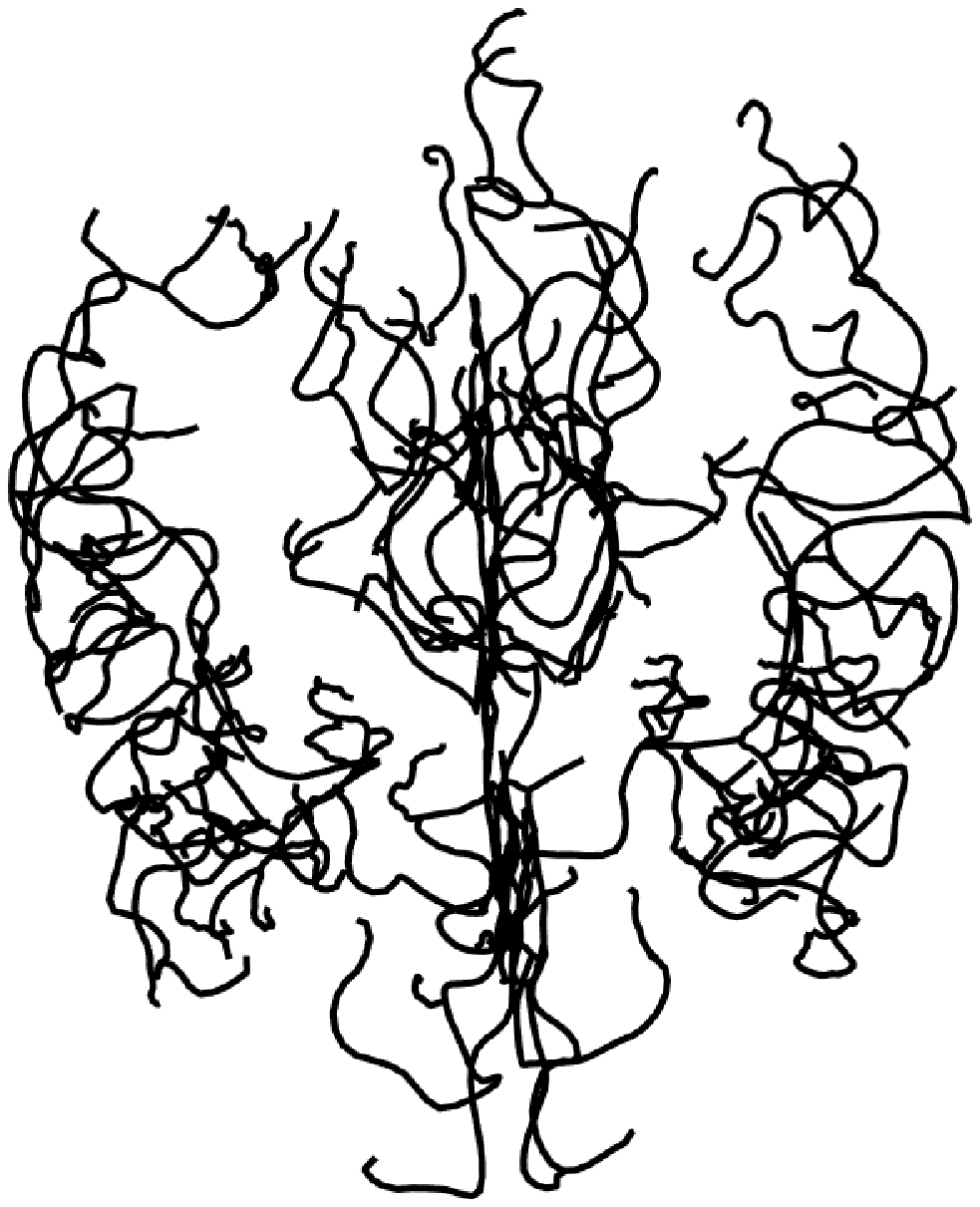}   
}
\caption{Some examples of graphs representing knowledge in different applications.}
\label{fig:sampleGraphs}
\end{figure}

Due to rapid advances in sensing and measurement technology, 
data is increasingly becoming complex and structured, reflecting
the growing need for newer approaches and problem formulations. 
One common approach to understanding complex, high-dimensional 
datasets is to represent
them as graphs. Typically one identifies several variables of interest in the data, 
designates them as nodes, and represents their interactions as edges. 
Such a graph captures variability and interactions associated with a large number of 
variables, and lends to higher-order statistical analysis. 
Examples of graphical representations can be found in many areas, including social networks~\cite{ugander2011anatomy}, 
gene expression networks \cite{westenberg2008interactive}, brain connectivity 
data \cite{dai2016testing}, geographical data~\cite{mackaness1993use},
financial stocks \cite{yang2003market}, communication networks \cite{hakimi1965optimum}, epidemiology \cite{shirley2005impacts}, and so on. 
Fig. \ref{fig:sampleGraphs} shows some examples:
a social network;
a molecule with atoms as nodes and chemical bonds as edges;
a video represented as a pattern theoretic graph \cite{de2014pattern}
with objects or actions as nodes and their relationships as edges; and a brain arterial graph whose edges are 2D/3D curves connecting nodes.  

We focus on the problems where one has several graphs, each representing a snapshot or an observation of a system. One is interested in capturing, modeling, and analyzing {\it statistical shape variability across these graphs}. For instance, consider the representation of functional connectivity of parts of a human brain during performance of a particular task, as measured by fMRI signals, using graphical structures. Given several of these graphs, 
one for each human subject under each task and performance, one has a large amount of graph data to analyze and model. Similarly, one may have graphical representations of different social or economic networks, each representing different communities. The
general goal of statistical analysis is: (1) derive {\it common characteristics across observed graphs}, 
(2) distinguish {\it graph-populations using statistical testing}, and (3) model variability in graph shapes
 using {\it analytical generative models}. 
Further, using these tools, 
one can generate synthetic graphs that follow dominant variability in observed complex graphs. An interesting use of generative models is in augmenting training data that enables
more data-driven deep learning approaches. 
In the long term, this theory can support deep learning on \textit{non-Euclidean spaces of graphs}, where the data elements are themselves graphs.

The structured nature of graphs makes them challenging to analyze using classical statistical
tools.  
A graph is a non-Euclidean data object consisting of a set of variables as nodes and their interactions as edges. 
There are two sources of variability in shapes of graphs -- (i) different number and values (attributes) of 
the nodes, and (ii)
different connectivity patterns and values of the edges.
Therefore, one is interested in mathematical representations that enable a quantitative statistical analysis of graph data using both edge and node attributes. 
For quantifying differences between graphs, one requires metrics that can 
incorporate measures of these properties. 
However, one big issue in analyzing shapes of graphs is that 
nodes across graphs often come without matchings or correspondences. 
Although it is possible to bypass the issue
in some subproblems, for example
by counting common substructures~\cite{kriege2020survey}, the problem of establishing
correspondences of nodes across 
graphs (named {\it registration or graph matching)} is fundamental and represents one of the 
biggest challenges in the statistical analysis of graphs. 

\subsection{Literature Review}
The use of graph representations is of great interest in several broad problem domains. We cover some of these areas in a brief survey.  

In machine learning, primarily geometric deep learning, the focus is on
learning geometry of the \textit{manifold of the data elements (nodes)}
which, in turn, can be naturally represented as a graph. 
In a related problem, some papers consider entire graphs as the entities of interest. For instance, 
papers, such as \texttt{graph2vec}~\cite{narayanan2017graph2vec} and \texttt{UgraphEmbed}~\cite{baiunsupervised}, 
consider the problem of assigning a vector space representation to entire graphs. 
Some other papers seek to find vector representations for nodes such that distances in the vector space preserve a graph's original neighborhood structure. 
Examples include modeling random-walks through the nodes~\cite{perozzi2014deepwalk,grover2016node2vec}, or ones that preserve first and second-order proximity information~\cite{wang2016structural,tang2015line}, or ones that consider larger neighborhood structures as captured by node coarsening~\cite{chen2018harp}.
Some other papers utilize graph convolutions, {\it i.e.}, convolutions over the nodes, respecting their neighborhood structure. 
Examples include graph signal processing~\cite{ortega2018graph,defferrard2016convolutional}, or works that direct processing via the local graph structure~\cite{kipf2016semi,hamilton2017inductive}. Similar to the embedding approaches, these papers also consider \textit{a} single graph and operate on the nodes of that graph. 
We refer the readers to \cite{sato2020survey} for an overview of Graph Neural Networks (GNN).

A central problem studied in this paper is graph matching. 
In this context, there are mainly two types of approaches: exact matching and inexact matching \cite{neuhaus2007bridging}.
The exact matching implies
finding a bijective map such that the nodes and edges across two graphs are in one-to-one correspondence.
If we can match two graphs exactly, then the mapping is also called an {\it isomorphism}.
A related topic is {\it subgraph isomorphism}~\cite{carletti2017challenging}, where one graph matches a subset of another graph. 
In contrast, inexact matching seeks optimal registration between graphs that may be dissimilar. 
The inexact matching is more common in practice because of the complexities associated with real data. 
Since matching of two sets of nodes is essentially a problem of combinatorics, the
problem of finding a global optimum for inexact graph matching is NP-complete \cite{conte2004thirty}. 
Therefore, most algorithms for graph matching seek approximate solutions based on different relaxations of the original problem. 
As described later, one achieves a mathematical representation of registration of nodes (across graphs) using a permutation group -- a permutation of the ordering of nodes in a graph changes its registration with an ordered set of nodes representing another graph. 
The approximate solutions result from expanding beyond the permutation group to some larger set where the solutions are more readily available. 
One idea is to replace  
permutations by orthogonal matrices and then use spectral 
(eigen-decomposition based) approaches to find optimal rotations, see \cite{umeyama1988eigendecomposition,caelli2004eigenspace}.
Another direction is to replace permutations with doubly stochastic matrices and find a solution in that larger space. 
In all these cases, the final solution is eventually restricted to the discrete set of permutation matrices, see e.g., \cite{gold1996graduated, lyzinski2016graph,zhou2015factorized}.
Besides these approximations, there are some other algorithms for approximate
graph matching \cite{almohamad1993linear,krcmar1994application,riesen2009approximate}.

There is also a significant amount of work on analyzing graphical data using graph kernel methods~\cite{kriege2020survey,shervashidze2011weisfeiler,vishwanathan2010graph,yanardag2015deep}.
The basic idea here is to design kernel functions that measure similarities between graphs. 
These kernels are then used in kernel-based methods, such as the support vector machine (SVM). 
There are also kernel-based statistical methodologies, including kernel mean embedding~\cite{song2013kernel} and kernel PCA~\cite{scholkopf1998nonlinear}.
However, the tasks performed by graph kernel methods are mainly discriminatory, and it is usually impossible to map the results from a feature space back to the original graph space.

\textcolor{black}{Recently, there has been progress in analysis of graphical data using optimal transport (OT)~\cite{chapel2020partial, chizat2018scaling, chowdhury2019gromov,memoli2011gromov,peyre2016gromov,sejourne2020unbalanced,vayer2018optimal}. } More specifically, these papers use Gromov-Wasserstein distance to handle the structured relational information.
The OT-based approaches seek a soft (probabilistic) registration of nodes, but we need a hard node permutation.
\textcolor{black}{Note that in certain cases, the soft registration is tight because the matching energy is concave and the solution is a permutation~\cite{maron2018probably}.}
The focus in our paper is on {\it shapes} of graphs, 
 both in a visual and a mathematical sense.
 In these situations, the geodesics and the mean shapes under the Gromov-Wasserstein framework with soft registrations will be messy because it will result in dense connections between nodes across graphs.
Furthermore, we seek a representation space of graphs where statistical tools, such as mean, covariance, and PCA, can be derived. So far, OT-based approaches have developed only some of these tools. 

The most relevant past research related to our approach comes from Jain et al. 
\cite{jain2009structure,jain2012learning,jain2016geometry,jain2016statistical}
who first introduced and developed a mathematical framework of representing graphs as elements of quotient spaces. These papers provided a rigorous theory for quotient space geometry and developed some basic statistical tools for graph data analysis.

\subsection{Our Contributions}
In this paper, we further develop the framework of 
\cite{jain2009structure,jain2012learning,jain2016geometry,jain2016statistical,kolaczyk2020averages},
leading to a comprehensive approach for comparing, summarizing, and analyzing the {\it shapes} of graphs. The basic idea, first introduced in \cite{jain2009structure}, represents graphs as matrices and formulates the registration problem as that of permutation of entries in those matrices. 
Mathematically speaking, we represent the registration variability using the permutation group's action on the set of matrices representing all graphs. To remove this nuisance group, 
we form a quotient space and inherit a metric on the quotient space from the original set
of matrices. This procedure is similar to the development of shape spaces in the statistical analysis of shapes~\cite{mardia-dryden-book,FDA}. 
We use a standard Euclidean metric with appropriate invariance properties because it allows for efficient registration of nodes across graphs. 
One can use the quotient space metric to define and compute statistical summaries, such as sample mean, covariance, and principal components. 
The principal component analysis or PCA helps perform dimension reduction and impose compact statistical models on observed graphs. These models play essential roles in hypothesis testing and other statistical inferences involving graph data.

\textcolor{black}{This paper borrows several ideas from the current literature and develops them into a more comprehensive theory that facilitates deeper statistical analysis and modeling.} 
The novel contributions of this paper are as follows: 
\begin{enumerate}
\item It adapts a quotient space metric structure on the set of graph representations,
originally introduced in \cite{jain2009structure} and 
further developed in 
\cite{jain2012learning,jain2016geometry,jain2016statistical,kolaczyk2020averages}, and extends it to include both node and 
edge attributes. It uses this metric structure to quantify graph 
differences and to compute optimal deformations (geodesics) between 
graphs. Using this metric structure, it establishes a framework for 
computing
sample statistics such as mean and covariance for graph data. 
\textcolor{black}{In comparison, although the paper \cite{jain2016geometry} includes both node and edge attributes
in the analysis, it relies on a kernel representation to do so. The limitations of a kernel-based approach have been 
noted earlier.}

\item A key idea here is that it does not assume the graphs to be
equal-sized and matched. That is, one allows nodes to remain unmatched across the graphs. Past metric-based 
approaches often insist on matching every node to a proper node during graph comparisons. 

\item It defines a notion of PCA for graphs and uses that
to develop low-dimensional representations of observed graphs. Unlike the previous work \cite{severn2019manifold}, the proposed PCA is invariant to the node ordering.

\item It develops a simple Gaussian-type model for capturing graph variability in observed
graphs and uses it to generate random samples from such graphical models. This sampling, 
in turn, can be used either to augment graph neural networks or Bayesian inferences involving graphical data, although we have not pursued that direction here. 
\end{enumerate}

The rest of this paper is as follows. 
Section~\ref{sec:represent} describes the chosen mathematical representation of graphs using symmetric matrices.
Section~\ref{sec:graphmatching} studies the graph matching problem using the action of the permutation group. 
Section~\ref{sec:node-and-edge}
extends this framework to include both node and edge attributes in the framework. 
Section~\ref{sec:statitics} presents techniques for statistical analysis of graph data. 
Section~\ref{sec:experiment} shows a number of experiments illustrating this framework.
The paper ends with a short discussion and conclusions in Section~\ref{sec:conclusion}. 
\section{Graph Representation and Metric Structure}
\label{sec:represent}
In this section, we will present a framework for the structure of graphs
first developed in \cite{jain2009structure,jain2012learning,jain2016geometry,jain2016statistical,kolaczyk2020averages}. 
We apply and advance this framework as described below. 

\subsection{Adjacency Matrix Representation}
We start by providing a mathematical representation for analyzing weighted graphs. 
A weighted graph $G$ is an ordered pair $(V,w)$, 
where $V$ is a set of nodes and $w$ is a weighting function: $w: V \times V \rightarrow \mathcal{M}$.
That is, $w(v_i, v_j)$ characterizes the edge between $v_i, v_j \in V, i \neq j$,
where elements of the set $E = \{(v_i,v_j) \in V \times V: i \neq j \}$ are the edges of $G$. 
In the literature, $\mathcal{M}$ is usually limited to be $\real^+$, i.e., non-negative real numbers. 
However, we allow $\mathcal{M}$ to be any Riemannian manifold on which one
can define distances, averages, and covariances. 
For example, 
in case of brain arterial network shown in Fig. \ref{fig:sampleGraphs},
the edges attributes are shapes of 3D curves connecting the nodes (junctions). 
Assuming that the number of nodes, denoted by $|V|$, 
is $n$, $G$ can be represented by its adjacency matrix 
$A = \{a_{ij}\} \in \mathcal{M}^{n \times n}$, where the element $a_{ij} = w(v_i, v_j)$. 
For an undirected graph $G$, we have $w(v_i, v_j) = w(v_j, v_i)$ and therefore $A$ is a symmetric matrix. 
(In this paper, we only focus on undirected graphs although the framework is extendable to 
directed graphs also.)
The set of all such matrices is given by ${\cal A} = \{ A \in \mathcal{M}^{n \times n} | A = A^T\}$. 

Let $d_m$ denote the Riemannian distance on $\mathcal{M}$. We will use this to impose a metric on the 
representation space ${\cal A}$. 
For example, $d_m$ can be 
the Euclidean distance if $\mathcal{M} = \real$ 
or a shape metric when $\mathcal{M}$ is the shape space.
For any two $A_1, A_2 \in {\cal A}$, 
with the corresponding entries $a_{ij}^1$ and $a_{ij}^2$, respectively, the 
metric $d_a(A_1, A_2) \equiv
\sqrt{\sum_{i,j} d_m(a_{ij}^1, a_{ij}^2)^2}$ quantifies the difference the graphs they represent. 
Under the chosen metric, the geodesic or the shortest path between two points in ${\cal A}$
can be written as a set of geodesics in $\mathcal{M}$ between the 
corresponding components. That is, for any $A_1, A_2 \in {\cal A}$, the geodesic $\theta: [0,1] \to {\cal A}$ 
consists of components $\theta = \{\theta_{ij} \}$ 
given by $\theta_{ij}: [0,1] \to \mathcal{M}$, a geodesic path in $\mathcal{M}$ between $\theta_{ij}^1$ and $\theta_{ij}^2$. 
In case $\mathcal{M} = \real$, then 
${\cal A}$ is a vector space, equivalent to a Euclidean space of dimension $n(n+1)/2$, 
and the geodesic between two points in ${\cal A}$
is a straight line. That is, for any $A_1, A_2 \in {\cal A}$, $\theta: [0,1] \to {\cal A}$ 
given by $\theta(t) = (1-t)A_1 + t A_2$ is the geodesic path. 

Since the ordering of nodes in graphs is often arbitrary, the subsequent analysis should not be dependent on this arbitrary choice. We view the ordering variability as a nuisance and seek to remove its influence from the analysis. 
A different way to state this issue is that nodes across graphs are registered during comparisons, and we will use permutations to perform registration. 
Let ${\cal P}$ be the set of all permutation matrices of size $n \times n$. A permutation matrix is a matrix that has exactly one 1 in each row and each column, with all the other entries being zero. 
This finite set forms a group, with the group operation being matrix multiplication and the identity element being the $n \times n$ identity matrix. 
Note that ${\cal P}$ is a subgroup of $O(n)$, the set of all $n \times n$ orthogonal matrices. 
For any $P \in {\cal P}$, the inverse of $P$ is given by $P^T$, the 
transpose of $P$. 
We define the action of ${\cal P}$ on ${\cal A}$ using the map: 
$$
{\cal P} \times {\cal A} \mapsto {\cal A},\ \ (P, A) = P A P^T \ .
$$
One can easily verify that this is a proper group action. 
For any $A \in {\cal A}$, its orbit under the action of ${\cal P}$ is given by: 
$$
[A] = \{ PA P^T | P \in {\cal P} \}\ .
$$
It is the set of all possible permutations of the node ordering in a graph represented by $A$.
Since ${\cal P}$ is a finite set, each orbit under ${\cal P}$ is also finite. 
Any two elements of an orbit denote the same graph, except that the ordering of the nodes has been changed. Therefore, the membership of an orbit defines an equivalent relationship $\sim$ 
on the set ${\cal A}$:
\begin{equation}\label{eq:equiva}
A_{1} \sim A_{2} \Leftrightarrow \exists P \in {\cal P}: P A_{1} P^T = A_{2}\ .
\end{equation}
One can check that any two orbits $[A_1]$ and $[A_2]$, for any $A_1, A_2 \in {\cal A}$, 
are either equal or disjoint. 
The set of all equivalence classes forms the quotient space or the {\it graph space}: 
\begin{equation}
{\cal G} \equiv {\cal A}/ {\cal P} = \{[A] | A \in {\cal A} \}\ .
\end{equation}
${\cal G}$ is a nonlinear space because it is a quotient space -- one cannot 
perform linear operations, such as addition or multiplications on its elements directly. 
For example, $x_1 [A_1] + x_2 [A_2]$ is not well defined in ${\cal G}$ for 
arbitrary $x_1, x_2 \in \real$. 
Actually, this type of space is called an {\it orbifold} \cite{jain2009structure,jain2012learning,jain2016geometry}. 
Next, we will impose a metric structure on this quotient space and use this metric to compute
statistical summaries and to perform statistical analysis. 

We can inherit the chosen distance $d_a$ from ${\cal A}$ 
on to the quotient space ${\cal G}$, but that requires the following result. 
\begin{lemma}
The action of $\mathcal{P}$ on ${\cal A}$ is by isometries under $d_a$. That is, 
for any $A_1, A_2 \in {\cal A}$ and $P \in {\cal P}$, we have 
\begin{equation}
d_a(P A_1 P^T , P A_2 P^T) = d_a(A_1 , A_2)\ .
\label{eq:isometry}
\end{equation}
\end{lemma}
The proof is easy since an identical permutation on both graphs leaves
the registration between nodes (across graphs) remains unchanged. 
This lemma enables the following definition. 
\begin{defn}[Graph Metric]
Define a metric on the graph space ${\cal G}$ according to: 
\begin{eqnarray}
d_g([A_1],[A_2]) &=& \min_{P \in {\cal P}} d_a(A_1, P A_2P^T) \nonumber \\
&=& \min_{P \in {\cal P}} d_a(A_2, P A_1P^T)
\label{eq:register}
\end{eqnarray}
\end{defn}
The last equality
comes from the fact that the action of ${\cal P}$ is by isometry (Eqn. \ref{eq:isometry})
and that ${\cal P}$ is a group. The minimizer $P^*$ provides the optimal registration
between graphs $A_1$ and $A_2$. That is, any element of $A_1$ is matched to 
the corresponding entry of the matrix $P^* A_2 P^{*T}$.

What can we say about the existence and uniqueness of the minimizer
in Eqn.~\ref{eq:register}?
Since ${\cal P}$ is finite, the minimum exists. The uniqueness, however, is not guaranteed in all cases. 
In theory, it is possible to have multiple permutations that attain the minimum of $d_a$. 
This can happen when an edge on one graph matches with more than one edge on the other graph. However, if the edge attributes are {\it continuous variables}, this event's probability is zero for non-zero edges. (It can happen for trivial or zero edges, but that can be characterized as trivial multiplicities). Therefore, in this case, the formulation enjoys the uniqueness of the registration solution as well. In the case of discrete edge attributes, there may potentially be multiple optimal registrations. In this case, any of these solutions work in the subsequent analysis.
This uniqueness issue is further studied experimentally later in Table~\ref{table1}.

One can define geodesics in the graph space ${\cal G}$ as follows. For any two graphs, 
with the adjacency matrices $A_1$ and $A_2$, and $P^*$ the optimal 
permutation of $A_2$ to best register it with $A_1$ (according to Eqn. \ref{eq:register}). 
Then, the geodesic path 
between $[A_1]$ and $[A_2]$ in ${\cal G}$ is given by the line 
$t \mapsto [\theta(t)]$, where the components $\theta_{ij}(t)$
denote geodesics in $\mathcal{M}$ between the registered elements of 
$A_1$ and $P^* A_2P^{*T}$. 
This geodesic, in turn, is useful in computing graph summaries and graph PCA, as defined later. For any two graphs with continuous-valued edge attributes, the registration solution exists and is unique
with probability one. Consequently, the corresponding geodesic between them also exists and is unique. There is a potential for multiple optimal registrations and multiple corresponding geodesics in the graphs with discrete edge attributes. We can select one of these multiple solutions for display and other uses. The lack of uniqueness of geodesics
is quite common in shape analysis and is often handled that way in practice.

\subsection{Alternative Representation: Laplacian Matrix}
In the special case when $\mathcal{M} = \real^+$, 
one can also use graph Laplacian matrix \cite{horaud2012short,severn2019manifold} as a
mathematical representation, instead of the adjacency matrix, for a graph. 
The graph Laplacian matrix $L = [l_{ij}]$ is defined as follows:
$$
l_{ij} = \left\{ \begin{array}{ll} 
-w(v_i,v_j), & \ \ \mbox{if}\ \ i \neq j \\
\sum_{k \neq i} w(v_i,v_k), & \ \ \mbox{if}\ \ i=j\ .
\end{array} \right. 
$$
(One can develop a notion for Laplacian on non-Euclidean domains also but with some additional 
geometric notation.)
One can consider Laplacian matrices to be elements of ${\cal L}$, the set of all positive semidefinite matrices of size $n \times n$. 
There is a bijective mapping between adjacency matrices and Laplacian matrices $\phi: {\cal A} \rightarrow {\cal L} $
with $\phi$ defined as follows. 
Suppose $A$ is an adjacency matrix and $L$ is the Laplacian matrix for the same graph $G$.
Then, $L = \phi(A) = D-A$, where $D =\text{diag}(A({\textbf 1}{\textbf 1}^{T}-I))$ and ${\textbf 1}$ is the vector of 
all ones.
The inverse of $\phi$ is given by: $\phi^{-1}: {\cal L} \rightarrow {\cal A}, A = \phi^{-1}(L) = \text{diag}(L)-L$.
The bijection of $\phi$ can be proved as follows. First, if $L_1=L_2$, $D_1-A_1=D_2-A_2$ and it implies $A_1=A_2$ (Injection). And $\forall L \in {\cal L}$, we can find the pre-image $A = \text{diag}(L)-L \in {\cal A}$ (Surjection).
There are some interesting properties associated with the two representations: 
\begin{enumerate}
\item Since $\text{diag}(PAP^T({\textbf 1}{\textbf 1}^{T}-I)) = P\text{diag}(A({\textbf 1}{\textbf 1}^{T}-I))P^T$, 
we have $\phi(P A P^T) = P \phi(A) P^T$, for all $P \in {\cal P}$.

\item For any geodesic path $\theta(t) = (1-t)A_1 + t A_2$ in ${\cal A}$, the corresponding path in ${\cal L}$ is 
given by: 
$\beta(t) = \phi(\theta(t))$
\begin{scriptsize}
\begin{eqnarray*} 
&& =(1-t)(\text{diag}(A_1({\textbf 1}{\textbf 1}^{T}-I))-A_1)+t(\text{diag}(A_2({\textbf 1}{\textbf 1}^{T}-I))-A_2) \\
&& =\text{diag}(((1-t)A_1+tA_2)({\textbf 1}{\textbf 1}^{T}-I)) - ((1-t)A_1+tA_2) \\
&& =(1-t)L_1+tL_2 \ .
\end{eqnarray*}
\end{scriptsize}
Note that $\beta(t)$ is generally not a geodesic path in ${\cal L}$ under the commonly used 
metrics on ${\cal L}$ (please refer to \cite{severn2019manifold} for these metrics). 
As an aside, we note that the adjacency matrices, in general, are not positive definite. So they are not elements of ${\cal L}$. 

\item A related fact is that, under the Frobenious norms on ${\cal A}$ and ${\cal L}$, the mapping $\phi$ is not isometric, 
i.e. $\| \phi(A_1) - \phi(A_2)\| \neq \| A_1 - A_2\|$, in general.
\end{enumerate}
The framework developed in this paper also applies to the Laplacian representation, 
instead of the adjacency representation. 
For simplicity, we mainly focus on the adjacency matrix representation in this paper. 

\section{Graph Matching Problem}
\label{sec:graphmatching}
The problem of optimizing over ${\cal P}$, as stated in Eqn. \ref{eq:register},
becomes a key step in evaluating the 
graph metric and performing statistical analysis. 
Let $G_1 = (V_1, w_1), G_2 = (V_2, w_2)$ be any two weighted graphs, 
and let $A_{1}, A_{2}$ be the corresponding adjacency matrices. 
To simplify the discussion on graph matching and existing literature, we will completely
focus on the case where $\mathcal{M} = \real$. 
(For the non-Euclidean domains, we refer the reader to a follow-up paper~\cite{guo-Brain:2020}.)
Then, 
the registration requires solving the problem:
\begin{equation}
P^* = \operatorname*{argmin}_{P \in {\cal P}} \|PA_{1}P^T - A_{2} \|^2\ .
\label{eq:Euc-opt}
\end{equation}

So far we have assumed that the two graphs being compared have the same number of nodes. 
In general,
the graphs $G_1$ and $G_2$ may have $n_1$ and $n_2$ nodes, respectively, with 
$n_1 \neq n_2$. To handle this situation, we
add $n_2, n_1$ {\it null nodes} to $G_1, G_2$, respectively, to bring each of them to the same size $n_1 + n_2$.
The null nodes are unattached nodes with zero values for the edge and node attributes. 
As a result, the new adjacency matrices of $G_1$ and $G_2$ are: 
\begin{equation}\label{eq:ext_mat}
A_{1}' = 
\begin{pmatrix} 
A_{1} & 0_{n_1 \times n_2}\\ 
0_{n_2 \times n_1} & 0_{n_2 \times n_2} 
\end{pmatrix},
A_{2}' = 
\begin{pmatrix} 
A_{2} & 0_{n_2 \times n_1}\\ 
0_{n_1 \times n_2} & 0_{n_1 \times n_1} 
\end{pmatrix}
\end{equation}
The new matrix dimensions are $A_{1}', A_{2}' \in \real^{(n_1+n_2) \times (n_1+n_2)}$ and, 
therefore, we are back to matching graphs of the same size. 
This idea of extending the adjacency matrix, using Eqn. \ref{eq:ext_mat}, can be applied even when the graphs being compared have the same number of nodes. This extension results in a more flexible matching since it allows the given nodes to match with null nodes. 
By doing this, one has more degrees of freedom in order to reach a better match. 
We elaborate on this flexibility later in the experiments section. 

In the next two subsections, we present two different solutions for 
the optimization problem stated in Eqn.~\ref{eq:Euc-opt}. 

\subsection{Umeyama Algorithm}
First, we introduce a classic solution from \cite{umeyama1988eigendecomposition} 
that is based on the eigendecomposition of representation matrices. 
This method is summarized in Algorithm \ref{algo:umeyama_ext} and not repeated in the text here. 
Note that Algorithm \ref{algo:umeyama_ext} applies to the current discussion with $\lambda = 0$, the more general case is discussed later in Section IV. 
As noted in \cite{umeyama1988eigendecomposition}, 
the solution $P$ is the global solution for isomorphic graphs
but is usually a good initialization for more general graph matching problems.
Thus, we use it as an initial condition for a greedy search (pairwise exchanges of rows and columns)
that seeks to improve the solutions further. 

\begin{figure}[t]
\begin{center}
\includegraphics[width=1.\linewidth]{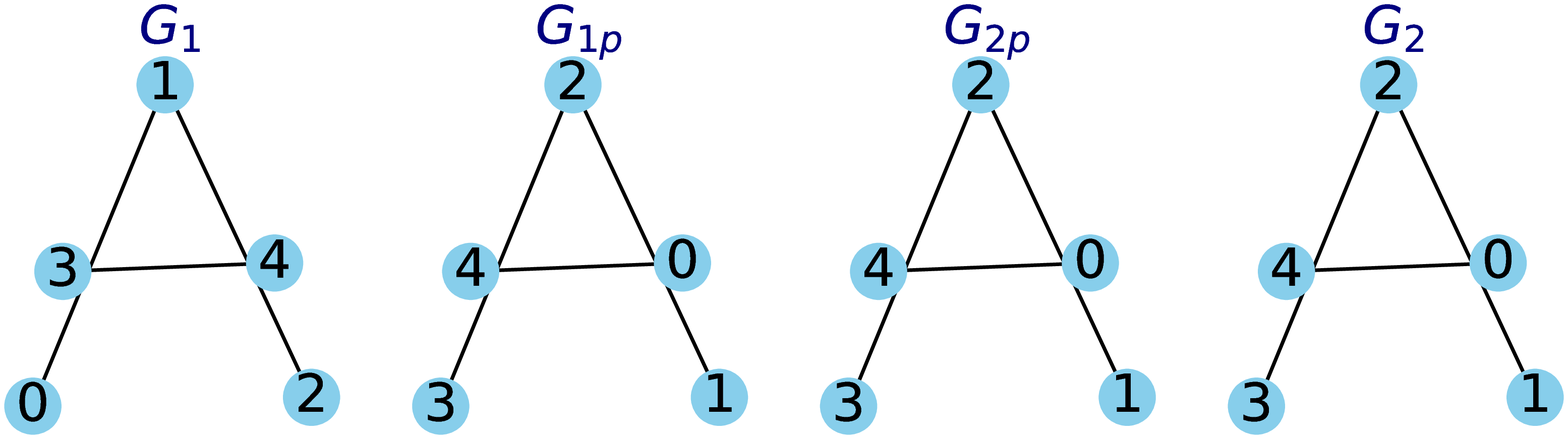}\\
\includegraphics[width=1.\linewidth]{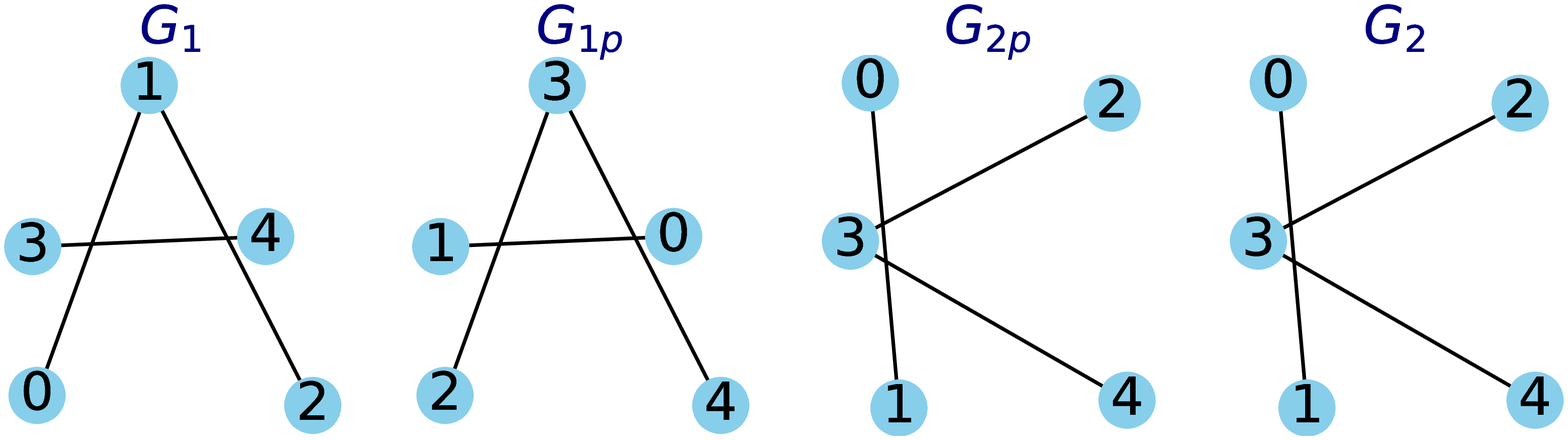}\\
\includegraphics[width=1.\linewidth]{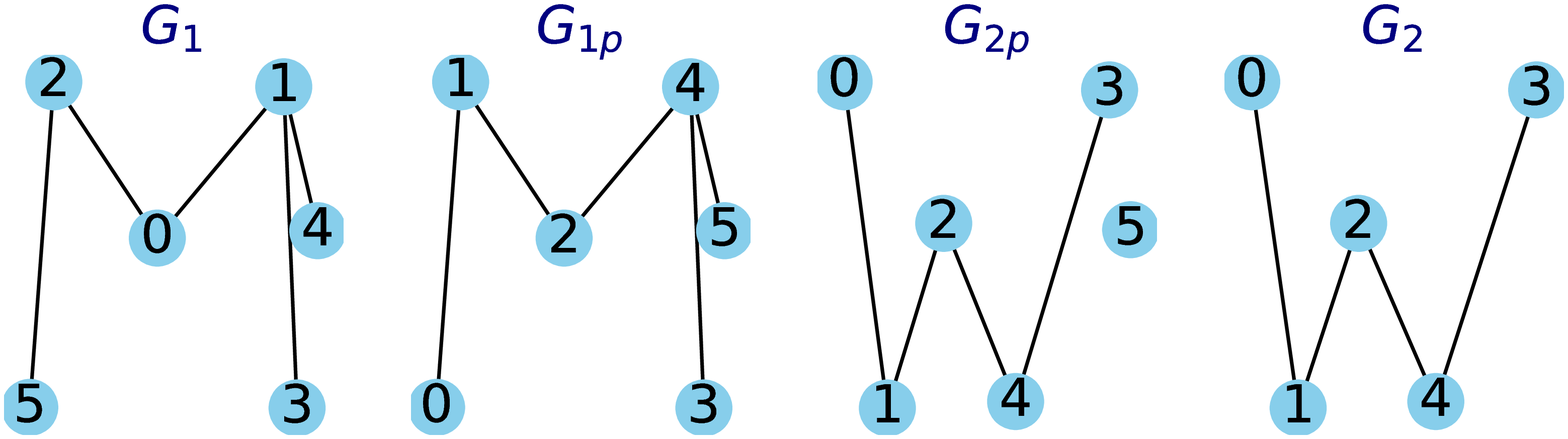}\\
\end{center}
\caption{Examples of graph matching using edge weights. In each row, the corner graphs, labeled $G_1$ and $G_2$, are the original graphs. The inner two, $G_{1p}$ and $G_{2p}$, are outcomes that are matched to each other. The outcome graphs may have some null nodes added, and the indices of $G_1$ are permuted as $G_{1p}$.}
\label{fig:edgeMatching}
\end{figure}

We illustrate this idea using some simple examples in Fig. \ref{fig:edgeMatching}. 
This dataset has binary graphs representing uppercase English letters~\cite{riesen2008iam}.
The edge attributes are binary in this example. Each row shows the original graphs 
$G_1$ (first graph) and $G_2$ (last graph), and their matched versions $G_{1p}$ and $G_{2p}$, in the middle.
$G_{1p}$ is the optimal permutation from Algorithm \ref{algo:umeyama_ext} of $G_1$, while $G_{2p}$ 
is same as $G_2$ with possibly some null nodes added. 
The first row shows the simpler case, 
where $G_1$ and $G_2$ have same number of nodes. 
We still add null nodes to both of them and permute $G_1$ to match $G_2$, resulting in $G_{1p}$.
As expected, the null nodes of $G_1$ are registered to null nodes of $G_2$, and are not displayed here.  
For the second row, the graphs $G_1$ and $G_2$ are different letters. 
In the last row, where the two graphs $G_1$ and $G_2$ have different sizes, a null node ${\it 5}$ has been added to reach a natural matching. For display purposes, we need to choose node attributes, {\it i.e.}, the placements of the null nodes. 
We have placed the null nodes at the same coordinates as their corresponding non-null nodes to improve the display.

\subsection{Fast Approximate Quadratic Programming}
Algorithm \ref{algo:umeyama_ext} generally works well for smaller graphs, but it slows down when the number of nodes gets large.
Recently, \cite{vogelstein2015fast, lyzinski2016graph} have used
the Frank-Wolfe algorithm \cite{frank1956algorithm} to develop a different solution, called {\it Fast Approximate Quadratic} or FAQ. 
The main idea is to restate matching problem according to: 
\begin{equation}\nonumber
\min_{P \in {\cal P}} \|PA_{1}P^T - A_{2} \|^2\ = \min_{P \in {\cal P}} \left( -\mathrm{Tr}(A_2PA_1P^T) \right)\ .
\end{equation}
The right-hand side of above equation is a special case of quadratic assignment problem.
One can solve it using the gradient of the cost function $f(P) = -\mathrm{Tr}(A_2PA_1P^T)$.
In order to handle the discrete nature of permutation matrix, the procedure 
first replaces the permutation matrix by a doubly stochastic matrix: 
\begin{equation}\label{eq:doublystochastic}
\min_{P \in {\cal D}} f(P) =\min_{P \in {\cal D}} \left( -\mathrm{Tr}(A_2PA_1P^T) \right) \ , 
\end{equation}
where ${\cal D}$ is the set of doubly stochastic matrices. 
These are matrices whose: (1) all entries are non-negative, and (2) rows sum, columns sum equal to one. 
After the optimization, the solution $P$ is projected back to the space ${\cal P}$. 
We summarize this approach in Algorithm \ref{algo:faq_ext}, with the current context applicable for 
$\lambda = 0$.

In theory, the algorithm should be able to find the exact matching when two graphs are isometric, even when we add null nodes to the graphs. 
However, how well does it perform in practice? 
To study this question, 
we randomly simulated graphs of
different types and sizes, and then randomly permuted their nodes.
We then solved for the registration between the original graph and the permuted 
graph using the FAQ method. 
(Although the two graphs are of the same size, we still add null nodes.)
Two types of graphs are studied -- binomial and fully-connected real-valued graphs. Table \ref{table1} 
below shows
the fraction of correct registrations measured over 1000 random graph matchings (for each entry). 
One can see that for continuous-valued edge attributes, the algorithm finds the optimal matching perfectly. However, in the case of binary graphs, especially with a small number of nodes, there is a 
possibility of multiple global solutions. Thus, the algorithm may not find the original registration. In all cases, the algorithm is very robust to the addition of null nodes. 

\begin{table}[h]
\begin{center}
\caption{Fraction of correctly registered nodes across random graphs for different graph types
and graph sizes.}\label{table1}
\begin{tabular}{|c|c|c|}
\hline
Number of Nodes & Binomial Graphs & Fully Connected \\
& & Weighted Graphs \\
& & (weights $\sim$ t(1)) \\
\hline
[5, 10] & 0.634 & 1 \\
\hline
[50, 60] & 0.992 & 1 \\
\hline
[100, 120] & 0.999 & 1 \\
\hline
\end{tabular}
\end{center}
\end{table}

\section{Extension Involving Both Edge Weights and Node Attributes}
\label{sec:node-and-edge}
In many cases, the structure of a graph can be identified by comparing edge weight exclusively. 
However, sometimes the information associated with the nodes of graphs is also crucial in matching 
and comparing graphs. 
Next, we extend the previous framework to incorporate node information also.  

Let $\mathcal{N}$ be 
the set of potential node attributes and 
let $\epsilon \in \mathcal{N}$ be a distinguished element denoting the null or void element.
A node-attributed weighted graph is represented by 
$G = (V, w, \alpha)$, 
consisting of: (i) a finite nonempty set $V$ of nodes, 
(ii) a weight function $w$ for edges, and (iii) 
an attribute function for nodes given by $\alpha: V \rightarrow \mathcal{N}$. 
Let $d_{\alpha}$ be an appropriate distance in $\mathcal{N}$, 
$d_{\alpha}: \mathcal{N} \times \mathcal{N} \rightarrow \real^+$.

For any two graphs $G_1 = (V_1, w_1, \alpha_1)$ and $G_2 = (V_2, w_2, \alpha_2)$, 
each with $n$ nodes,
let $D$ denote the $n \times n$ matrix of pairwise 
squared distances between nodes across the two graphs.  
That is, 
$$
D = [d_{ij}=d(\alpha_1(v_i^{(1)}),\alpha_2(v_j^{(2)}))^2] \in \real ^{n \times n},
$$ 
where $v_i^{(1)} \in V_1, v_j^{(2)} \in V_2, i, j = 1,2,..., n$. 
Now the matching problem becomes:
\begin{equation}\label{eq:obj_nodes}
P = \operatorname*{argmin}_{P \in {\cal P}} \{ ||PA_{1}P^T - A_{2} ||^2 + \lambda \mathrm{Tr}(PD) \} \ ,
\end{equation}
where $\lambda>0$ is the tuning parameter to balance the contributions of edge and node 
attributes in matching.
Using the same arguments as for Eqn.~\ref{eq:register}
earlier, a solution for this matching problem exists and is also 
unique (with probability one) when the node and edge attributes are continuous variables.
For FAQ, the equivalent matching problem is defined as:
\begin{equation} \label{eq:faq_ext}
P = \operatorname*{argmin}_{P \in {\cal P}} \left[ -\mathrm{Tr}(A_2PA_1P^T) + \lambda \mathrm{Tr}(PD) \right]\ .
\end{equation}
The gradient for 
this objective function is $(-A_2PA^T_1-A^T_2PA_1 + \lambda D^T)$.
The previous algorithms can be simply modified to handle the 
new formulation. In fact the solutions of this extended problem are already presented in 
Algorithms \ref{algo:umeyama_ext} and \ref{algo:faq_ext} for a general $\lambda$.

More generally, for $G_1, G_2$ with $n_1, n_2$ ($n_1 \leq n_2$) nodes, we extend 
the $n_1 \times n_1$ matrix $D$ 
according to: 
\begin{equation}
D' = 
\begin{pmatrix} 
D & D^*_{n_1 \times n_2}\\ 
D^*_{n_2 \times n_1} & 0_{n_2 \times n_2} 
\end{pmatrix}\ .
\label{eq:null-node}
\end{equation}
Here, the off-diagonal  
elements $d_{ij} = d(\alpha_1(v_i^{(1)}),\epsilon)^2$ in $D^*_{n_1 \times n_2}$ represent 
the node-attribute squared distance between $v_i^{(1)} \in V_1$ and $j$th null node $\epsilon$
in $G_2$. 
The choice for the null node attribute is arbitrary.
The smaller values in $D^*_{n_1 \times n_2}$ encourage real nodes to register to null nodes while larger values prevent it. 
The explanation applies to $D^*_{n_2 \times n_1}$ as well.
\textcolor{black}{There are also some similar ideas used in the G-W framework for comparing graphs with different sizes~\cite{gramfort2015fast,pele2008linear}}

\begin{algorithm}
\caption{Umeyama with Extension Involving Node Attributes}\label{algo:umeyama_ext} 
Given graphs $G_1$ and $G_2$ and the associated adjacency matrices $A_{1}$
and $A_{2}$, and $D$ is the node attribute squared distance matrix.
\begin{algorithmic}[1]
\State Compute the eigendecompositions $A_{1} = U_{1}\Sigma_{1} U_{1}^T$ and $A_{2} = U_{2}\Sigma_{2} U_{2}^T$
\State Find $P = \operatorname*{argmax} \mathrm{Tr}(P^T (\bar{U}_{1}\bar{U}_{2}^T - \lambda D^T ))$ 
using the Hungarian algorithm \cite{kuhn1955hungarian}. 
As earlier, $\bar{U}_{i}, i =1,2$ denotes a matrix with values that are magnitudes of the 
corresponding 
elements of $U_{i}$.
\State (Optional) Find the best exchange of two nodes of $G_1$ based on $P$, call it $P^*$, such that $P^* = \operatorname*{argmin}_{}\|PA_{1}P^T - A_{2} \|^2+ \lambda \mathrm{Tr}(PD)$ and update $P = P^*$
\State Repeat 3 until the value of  
$\|PA_{1}P^T - A_{2} \|^2+ \lambda \mathrm{Tr}(PD)$ does not decreases.
\end{algorithmic}
\end{algorithm}

\begin{algorithm}
\caption{FAQ with Extension Involving Node Attributes}\label{algo:faq_ext} 
Given graphs $G_1$ and $G_2$ and the associated adjacency matrices $A_{1}$
and $A_{2}$, and $D$ is the node attribute squared distance matrix.
\begin{algorithmic}[1]
\State Choose an initial $P \in {\cal P}$.
\State Compute the gradient of $f(P)$: $\nabla f(P) = -A_2PA^T_1-A^T_2PA_1+\lambda D^T$.
\State Approximate $f(P)$ by first order Taylor expansion around the current estimate $P^*$: $f(P) \approx f(P^*)+\nabla f(P^*)^T(P-P^*)$ and use Hungarian algorithm to minimize it, get $Q$.
\State Line search to determine the optimal step size $\eta \in (0,1)$
\State Update the doubly stochastic matrix $P^* = P^* + \eta(Q-P^*)$
\State Repeat 2-5 until convergence
\State Project back to the permutation matrix using Hungarian algorithm.
\end{algorithmic}
\end{algorithm}

Our approach here for choosing the null node attribute $\epsilon$ is the following. We let this value be a variable and 
let the data dictate what it should be. 
Note that the node attributes are entered in the framework through the extended distance matrix $D^*$, 
so the choice of node attributes is implicit. In Eqn.~\ref{eq:null-node}
we set all entries of $D^*$ corresponding to one or both null nodes to be zero. This implies that the null node attribute is set equal to the attribute of the matched node. 
So, the null node attribute can change from node to node and registration to registration, but this attribute is never explicitly stated. This choice is logical because the introduction of null nodes does not contribute to the cost. In other words, we do not want the attributes of the null node (which is a synthetic addition) to add to the matching cost in any way.

\begin{figure} 
\centering
\subfloat[$\lambda = 0$]
{
\includegraphics[width=1\linewidth]{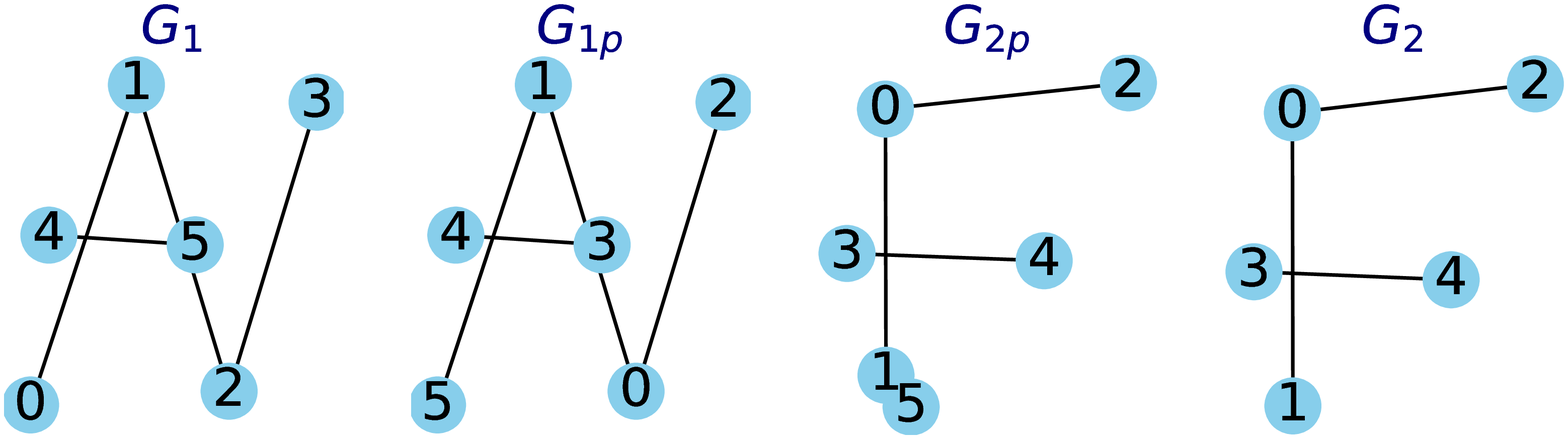}   
}\\
\subfloat[$\lambda = 0.5$]
{
\includegraphics[width=1\linewidth]{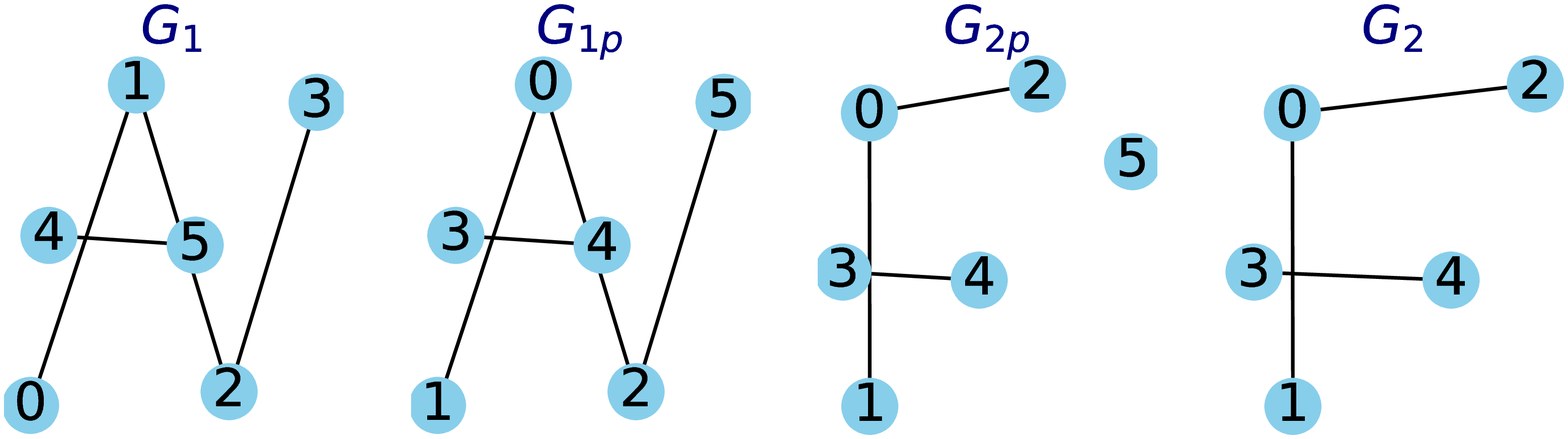}   
}\\
\subfloat[$\lambda = 1$]
{
\includegraphics[width=1\linewidth]{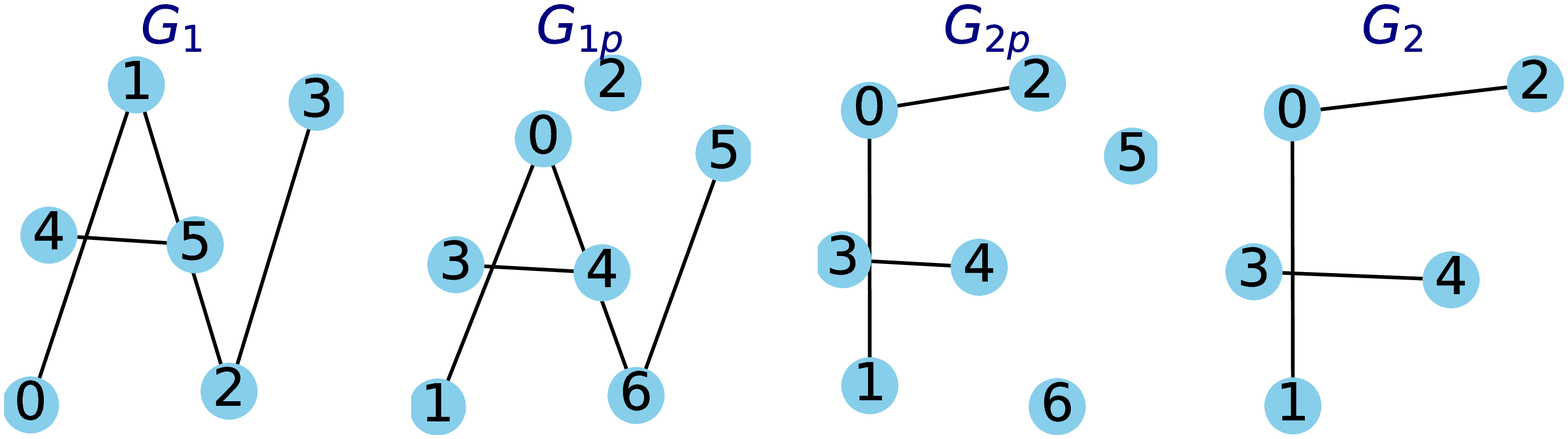}   
}

\caption{Example of graph matching using both edge weight and node attributes with different $\lambda$. In each row, the outermost graphs, labeled $G_1$ and $G_2$, are the original graphs. The inner two, $G_{1p}$ and $G_{2p}$, are the 
matched graphs. The matched graphs may have some 
null nodes added, and the indices of $G_1$ are optimally permuted to reach $G_{1p}$.}
\label{fig:edgeNodeMatching}
\end{figure}

In Fig. \ref{fig:edgeNodeMatching} we present some illustrations 
of graph matching using both edge and node attributes.
In this example, we use the planar coordinates of nodes of letter graphs as their attributes and keep the edge attributes binary-valued. 
The first row is the case without using any node attributes, i.e. $\lambda = 0$ (in Eqn. \ref{eq:obj_nodes}). 
In second row, we add node attributes with $\lambda = 0.5$.
Compared to the first row, this case shows a better correspondence across graphs since the edges ${\it 0-1}$ are now registered across graphs.
If we further increase the node attributes' weight, as the last row ($\lambda=1$) shows, the matching mostly ignores the edge correspondence. 
In the second row, only one edge ${\it 2-5}$ of $G_{1p}$ is matched to a null edge,
while in the last row, two edges: ${\it 0-6}$ and ${\it 5-6}$ of $G_{1p}$ are matched with null edges.
We can also see both $G_1p$ and $G_2p$ are padded with null nodes in the last row. 

\begin{figure*} 
\centering
\subfloat[Geodesic Interpolation in ${\cal A}$]
{
\includegraphics[width=1\linewidth]{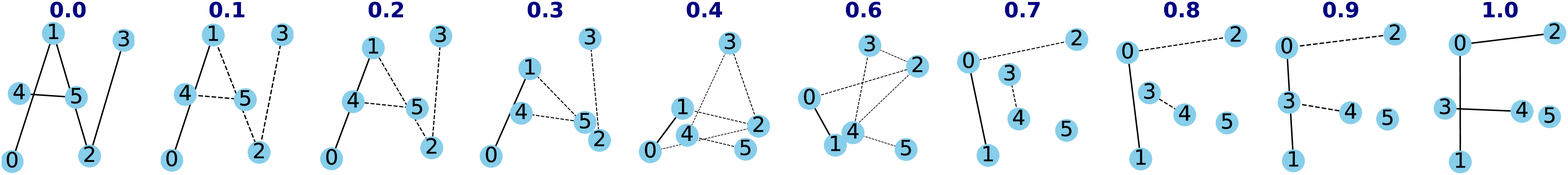}  
}\\
\subfloat[Geodesic Interpolation in ${\cal G}$]
{
\includegraphics[width=1\linewidth]{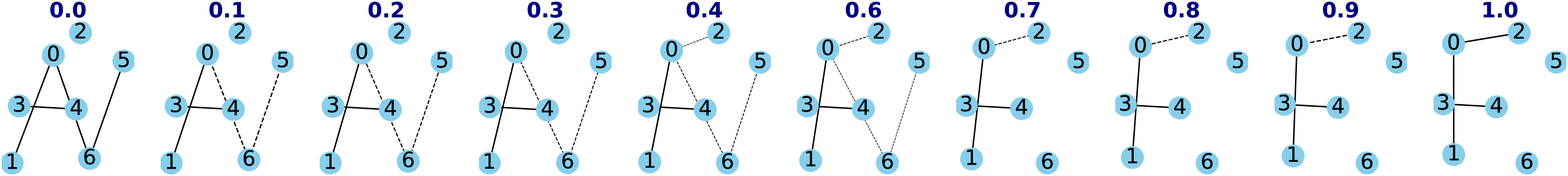}  
}
\caption{Comparison between geodesics in original space and graph space for two different graphs, $\lambda = 1$. The time point is labeled on the top of each graph, while 0 and 1 indicate the original graphs. Dash lines imply that the edges are changing. }\
\label{fig:geodesic}
\end{figure*}

An important strength of this framework is that
it provides geodesic paths between graphs in the quotient space ${\cal G}$. The geodesics in the pre-space ${\cal A}$ and the graph space ${\cal G}$ are linear interpolations, except for the optimal registration in the latter case. 
Fig. \ref{fig:geodesic} is a comparison between 
geodesics in ${\cal A}$ (top row) and ${\cal G}$ (bottom row) between 
the same two graphs. 
The two original graphs are at the two ends, representing the letter 'A' and the letter 'F'. 
In this example, we also use the 2D (centered) coordinates of nodes as 
node attributes with $\lambda = 1$. 
As one can see, geodesic in ${\cal G}$ shows a more natural 
{\it deformation} from one graph to the other, resulting from an improved matching of nodes. 

\begin{figure} 
\centering
\subfloat[Geodesic Interpolation in ${\cal G}$ (Adjacency Representation)]
{
\includegraphics[width=1\linewidth]{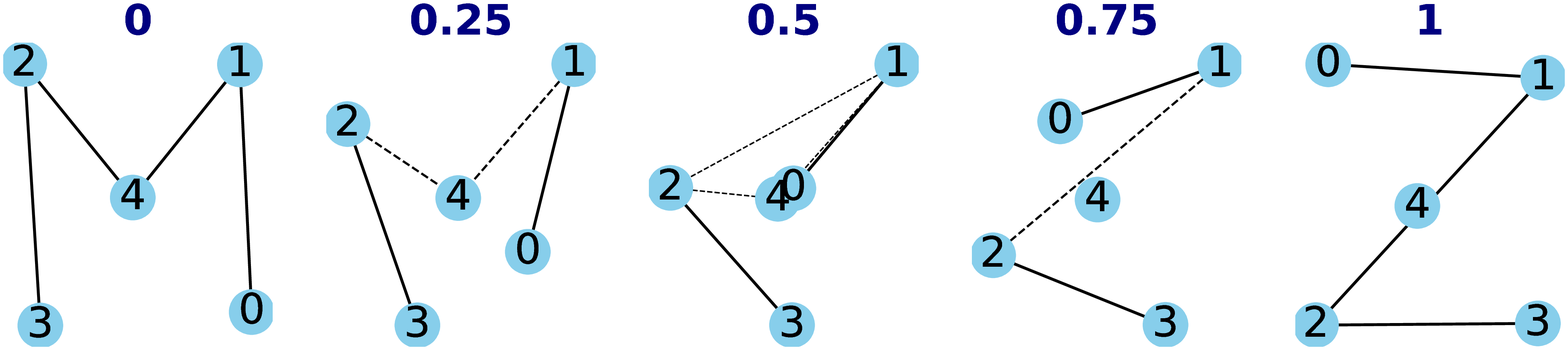}  
}\\
\subfloat[Geodesic Interpolation in ${\cal G}$ (Laplacian Representation)]
{
\includegraphics[width=1\linewidth]{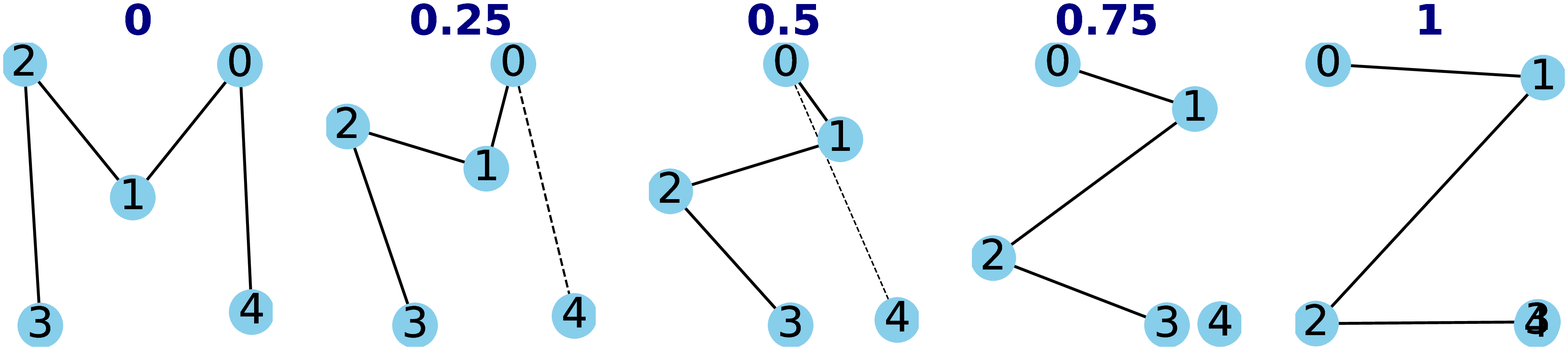}  
}
\caption{Comparison between geodesics in graph space using adjacency 
 and Laplacian matrix representation, $\lambda = 0$. Time point is labeled on the top of each graph while 0 and 1 indicate the original graphs. Dash lines 
imply that the edges are changing. }\
\label{fig:geodesicAL}
\end{figure}

As stated in the previous section, we can also use Laplacian matrices to represent graphs.
Although one can easily map an adjacency matrix to a Laplacian matrix using $\phi$, and vice-versa, 
the past literature has rarely used the Laplacian representation for graph matching. 
We present an example in Fig. \ref{fig:geodesicAL} where we perform matching under both the representations -- adjacency matrix and Laplacian matrix, using Algorithm \ref{algo:umeyama_ext}.
This implies that we are using the Frobenious norm for the Laplacian representation, 
although other norms can also be used instead. 
Since the mapping $\phi$ is not an isometry under the Frobenious norm between the
two representations, the minimization of $\|PA_{1}P^T - A_{2} \|^2$  
results in a different solution than minimization of $ \|PL_{1}P^T - L_{2} \|^2$.

To demonstrate the generalizability of this framework to more complex (edge and node) attributes, we present an example of registration and geodesic involving {\it shape graphs}. Here the edges are 2D curves connecting some planar nodes, and their attributes are shapes of these curves. The shape space of these curves is non-Euclidean and requires more elaborate procedures for computing
node registration and graph geodesics. For more details on these constructions, we refer the reader to the paper~\cite{guo-Brain:2020}. 
As the figure shows, the geodesic in $\mathcal{G}$ shows a
more natural deformation than that in $\mathcal{A}$.

\begin{figure} 
\centering
\subfloat[Geodesic Interpolation in ${\cal A}$]
{
\includegraphics[width=1\linewidth]{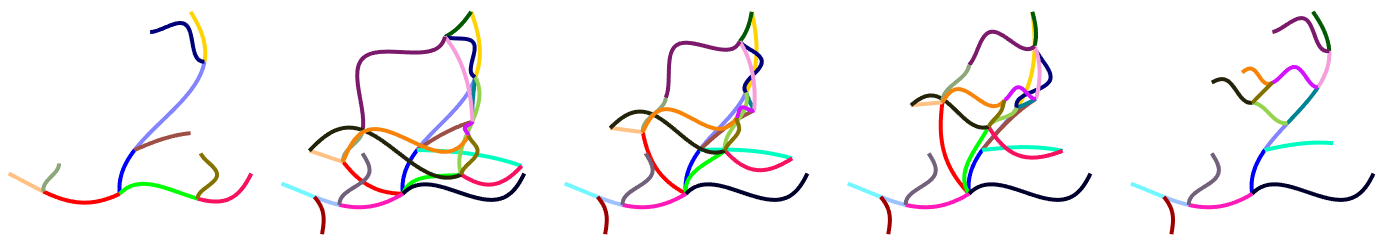}  
}\\
\subfloat[Geodesic Interpolation in ${\cal G}$]
{
\includegraphics[width=1\linewidth]{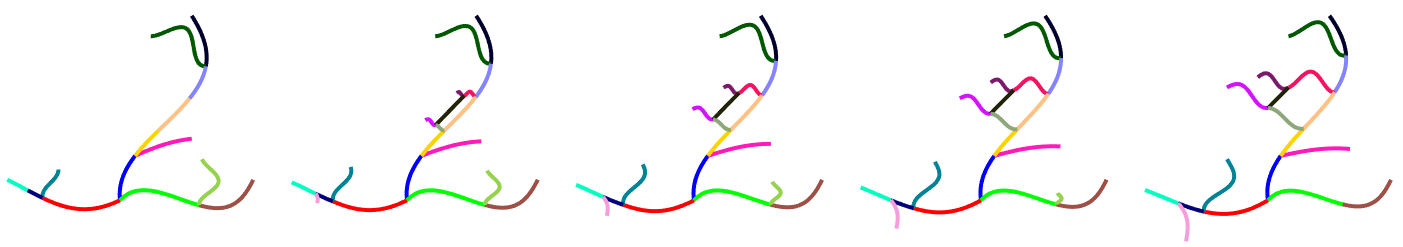}  
}
\caption{Comparison between geodesics in original space and graph space for two different graphs when $\mathcal{M}$ is a shape space. The leftmost and rightmost are two graphs whose edges are attributed as shapes. The correspondence is encoded in color.}\
\label{fig:geodesic_eg}
\end{figure}

\begin{figure}
\begin{center}
\hspace*{-0.6in} \includegraphics[height=0.9in]{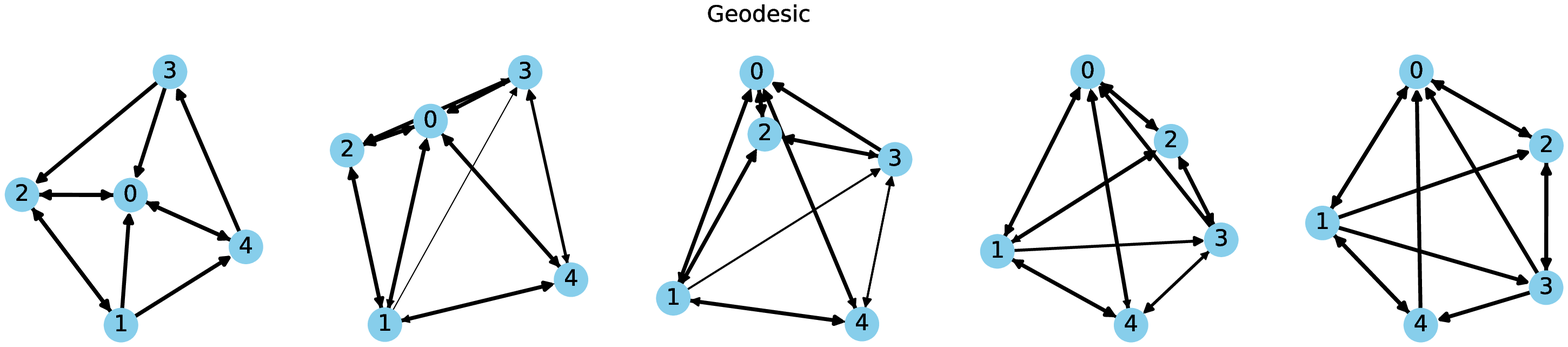} \\
\hspace*{-0.6in} \includegraphics[height=0.9in]{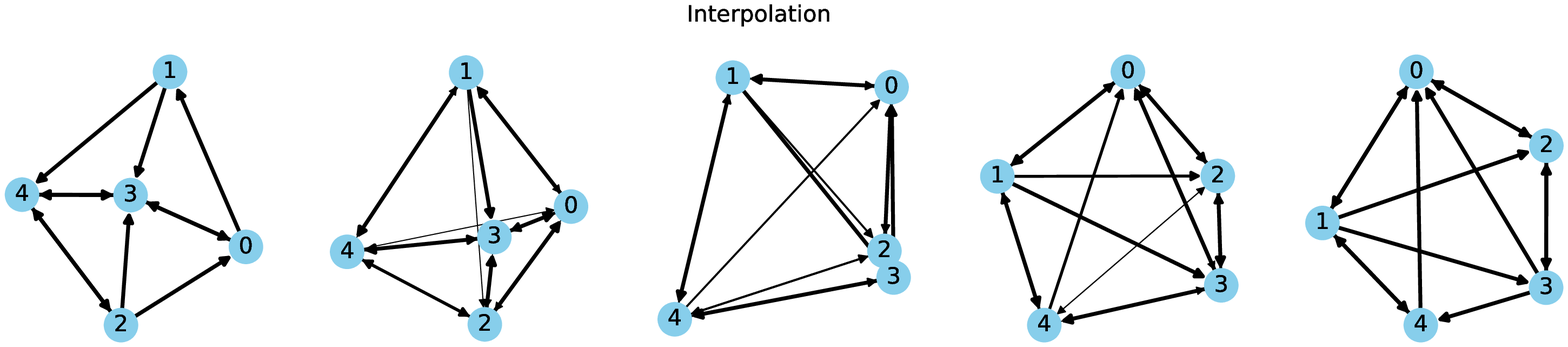} \\
\end{center}
\caption{An example of comparing two directed graphs: the top shows the geodesic in 
the quotient space, while the bottom shows an interpolation in ${\cal A}$.} 
\label{fig:directed}
\end{figure}

Another simple extension of this framework is to include directed graphs, 
{\it i.e.}, have graphs with directed edges. 
The only difference between directed and undirected graphs is that the symmetry of the adjacency matrices
is lost for the directed graphs. 
All the other parts of the approach remain the same. 
We present an example 
 between geodesic in graph space and interpolation in adjacency matrix space 
 in Fig.~\ref{fig:directed}. The geodesic in graph space better matches the edges and thus shows a 
 more natural deformation. 

\section{Statistical Analysis of Graphs}
\label{sec:statitics}
We have developed a metric space ${\cal G}$ for representing, matching, and comparing graphs. 
Additionally, we have tools for computing geodesic paths in ${\cal G}$ between arbitrary graphs. Together, these tools help us derive statistical summaries of graph data and develop analytical stochastic
models to capture the observed variability in given data. We start by defining sample means and covariances. 

\subsection{Mean of Graph Data}
Given a set of graphical data, it is important to 
summarize given graphs using the notion of a mean or a median. 
However, a simple average of the adjacency matrices does not make much sense if the 
nodes are not registered, which is usually the case in practice. 
Therefore, we would like to seek the mean in the graph space ${\cal G}$
\cite{jain2009structure,jain2012learning,jain2016geometry,jain2016statistical}.  
Given a set of $m$ graphs, $G_i \in {\cal G}$, 
$i = 1, .., m$, with corresponding adjacency matrices 
$A_{i} \in \real^{n \times n}$, the adjacency matrix of the mean graph is defined as:
\begin{equation}
[A_{\mu}] = \operatorname*{argmin}_{A \in {\cal G}} \sum_{i=1}^{m} d_g([A], [A_{i}])^2 \ ,
\label{eq:mean-A}
\end{equation}
where $d_g([A], [A_{i}])$ is as defined in Eqn. \ref{eq:register}.

What about the existence and uniqueness of this mean graph?
As discussed earlier, the optimal registration between any two graphs (with continuous-valued 
attributes) exists and is unique. That is, for a general $A$, the optimal permutations
$A_i^*$ of $A_i$ that attain $d_g([A], [A_i]) = d_a(A, A_i^*)$ are unique. Furthermore, 
given all $A_i^*$, their Euclidean mean is also unique, resulting in the existence and uniqueness of $[A_{\mu}]$. For graphs with discrete attributes, the uniqueness of registration is not
guaranteed, and the mean $[A_{\mu}]$ may be set-valued. That is, there may be several different graphs that attain the minima in Eqn.~\ref{eq:mean-A}. In that case, one can use any element of this set for further analysis.

An algorithm for computing this mean is given next. This algorithm is based on a greedy algorithm and may result in a local solution of the optimization problem presented in Eqn.~\ref{eq:mean-A}.
One can use improved initializations, 
stochastic searches, or some more advanced ideas to mitigate this problem.
\begin{algorithm}
\caption{graph Mean in ${\cal G}$}
\label{algo:mean}
Given graphs $G_i$ and the associated adjacency matrix $A_{i}$, $i=1,..,m$:
\begin{algorithmic}[1]
\State Initialize a mean template $A_{\mu}$ (e.g., the largest one).
\State Match $A_{i}$ to $A_{\mu}$ using Algorithm \ref{algo:umeyama_ext} or \ref{algo:faq_ext}
and store the matched graph as $A_{i}^*$, for $i = 1,.., m$.
\State Update $A_{\mu} = \frac{1}{m}\sum_{i=1}^{m} A_{i}^*$. In case we include 
node attributes in the analysis, we also perform an averaging of the registered nodes, as discussed below. 
\State Repeat 2 and 3 until $\sum_{i=1}^{m} \|A_{i}^*-A_{\mu}\|^2$ convergence.
\end{algorithmic}
\end{algorithm}

In case the node attributes are also included in the analysis, we will need
to endow the node attribute space ${\cal N}$ with a metric structure so that one can 
average the node attributes directly. 
For Euclidean node attributes, that is straightforward. However, in the case of 
categorical node attributes, one needs to embed these values in a metric space (such as 
one-hot encoding) and 
use that structure to compute the mean node value.
We observe that when comparing multiple graphs, the two-way null nodes padding 
(making both graphs the same size $n_1 + n_2$) results in a slower algorithm. 
Therefore, in some cases, we implement a faster approximation that applies only one-way null node padding.
Given multiple graphs, we pad all the other graphs with null nodes to register with the largest graph.

\subsection{Principal Component Analysis (PCA) of Graphs}
The high dimensionality of graph data is a problem in many domains. 
For a graph with $n$ nodes, the number of potential edges
can be as high as $\binom{n}{2}$. It will be useful to have a technique for projecting graph data to smaller dimensions while capturing as much intrinsic variability in the data as possible. 
We use PCA as a simple linear projection and dimension reduction tool and discover dominant directions/subspaces in data space.
As mentioned earlier, the non-registration of nodes in the raw data can be an obstacle in applying PCA directly on elements of ${\cal A}$. 
Instead, one can apply PCA in the quotient space ${\cal G}$, as summarized in Algorithm \ref{algo:PCA}. 
With PCA, graphs are represented as low-dimensional 
Euclidean vectors that facilities further statistical modeling and analysis. 
While several graph embedding techniques can be used to map graphs into Euclidean vectors, 
see for example \cite{bahonar2019graph}, the PCA approach comes with well-known optimality, and one can map the principal scores back to the graphs.

Given a set of graphs with adjacency 
matrices $A_i \in {\cal A}$, let $[A_{\mu}]$ denote their sample mean in ${\cal G}$
(obtained using Algorithm \ref{algo:mean}) and 
$A_i^*$ be the matrices registered to $A_{\mu}$. Then, the differences 
$\{ A_i^* - A_{\mu}\}$ are elements of the vector space ${\cal A}$ 
and \textcolor{black}{one can vectorize them and compute a covariance matrix, whose SVD 
provides the desired PCA}. The algorithm for PCA 
follows. 

\begin{algorithm}
\caption{Graph PCA}
\label{algo:PCA}
Given graphs $G_i$ and the associated adjacency matrix $A_{i}$, $i=1,..,m$:
\begin{algorithmic}[1]
\item Obtain the mean $A_{\mu}$ and the matched graph $A_i^*, i = 1,2,..,m$, using Algorithm \ref{algo:mean} 
\item Vectorize $A_i^* - A_{\mu}$, \textcolor{black}{compute their covariance matrix and use 
SVD of this matrix to perform PCA}. 
Obtain directions and singular values for the principal components. 
\end{algorithmic}
\end{algorithm}

The extended adjacency matrices can be used for the graphs with the different sizes, as mentioned before. 
One can also append $\{ A_i^* - A_{\mu}\}$ with node attributes when nodes are taken into account.
The details are left out here to save space. 

\subsection{Generative Graph Model}
In some situations involving statistical inferences, it is useful to develop analytical generative models for graphical data.
For example, it can be useful to produce new chemical molecules \cite{white2010generative}.
However, model estimation directly from observed graphs may have extraneous variance because the graphs are not registered. 
We introduce a simple Gaussian-type model in graph space ${\cal G}$ to better capture the essential variability of graphical data. 
In conjunction with the graph PCA and potential dimension reduction, we can reach a straightforward and efficient statistical model. 

Assume that we have a set of graphs with adjacency matrices $A_i, i = 1, ..., m$. 
By applying Algorithm \ref{algo:PCA}, we can get the PC scores $s_i \in \real^k$ by projecting each $A_i$ to the first $k$ principal components space.
For modeling the principal scores $s_i$'s, we impose a $k$ dimensional Gaussian model with sample mean and covariance as the model parameters. 
Note that while we use a Gaussian model here, one can also use any other parametric or nonparametric model instead. 
As alternatives, some machine learning methods \cite{sato2020survey} can also
provide ways of reaching Euclidean representations of graphs and learning generative models directly. 
For example, \cite{han2015generative} presents a generative graph model where one needs to gradually add nodes and edges to get a new sample graph.

\section{Experiments and Applications}
\label{sec:experiment}

To illustrate this framework, we have implemented it on various graph datasets and present the results next.

\subsection{Letter Graphs}
\begin{figure} 
\centering
\subfloat[Low Distortion]
{
\includegraphics[width=1\linewidth]{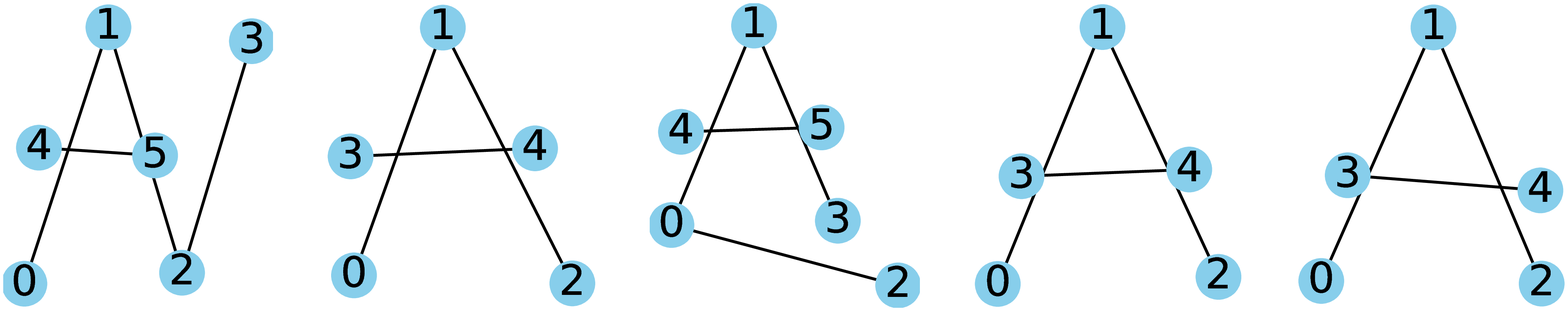}   
}\\
\subfloat[Medium Distortion]
{
\includegraphics[width=1\linewidth]{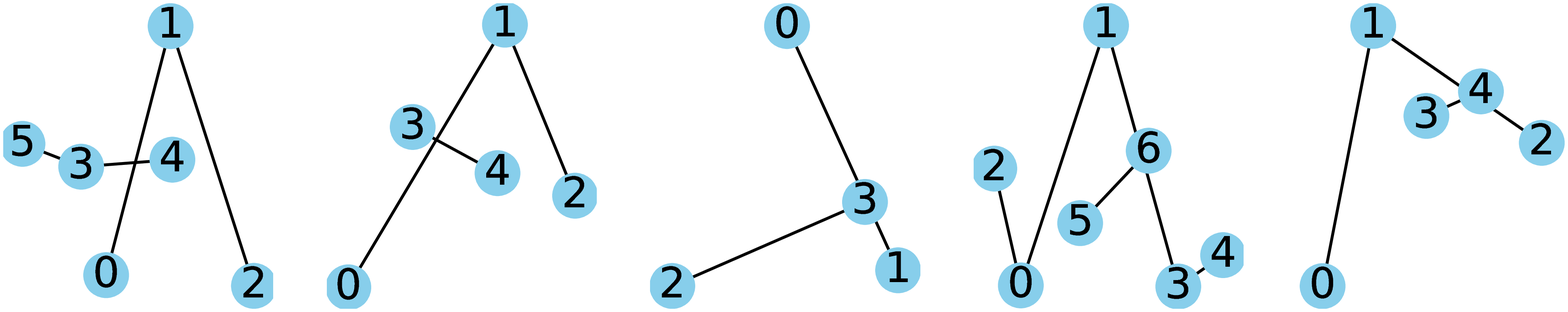}   
}\\
\subfloat[High Distortion]
{
\includegraphics[width=1\linewidth]{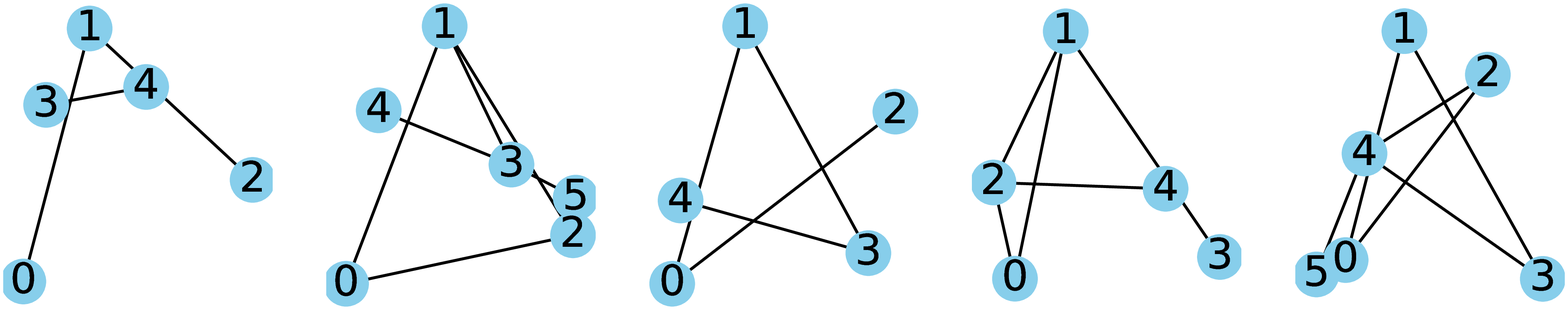}   
}
\caption{Sample graphs of Letter 'A' in different levels of distortion}
\label{fig:sampleA}
\end{figure}

The Letter Graphs dataset is a part of the IAM Graph Database used in \cite{riesen2008iam},
and consist of small graphs depicting 15 uppercase letters 
(A, E, F, H, I, K, L, M, N, T, V, W, X, Y, Z) that can be drawn using straight lines.
The edge weights in these graphs are either one or zero. 
The nodes have location coordinates in $\real^2$ so that the collection of edges form the shape of a letter.
The authors also introduced distortions to the prototype graphs at three different levels: low, medium, and high. 
Fig. \ref{fig:sampleA} shows some sample graphs of letter 'A' 
at these three different distortion levels. 

\begin{figure} 
\centering
\subfloat[Low Distortion]
{
\includegraphics[width=0.33\linewidth]{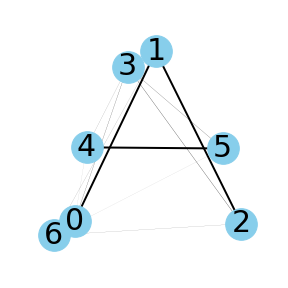}
}
\subfloat[Medium Distortion]
{
\includegraphics[width=0.33\linewidth]{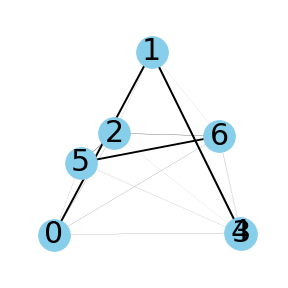}
}
\subfloat[High Distortion]
{
\includegraphics[width=0.33\linewidth]{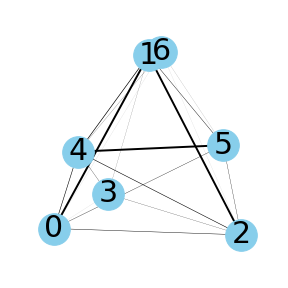} 
}
\caption{Mean graphs of letter 'A'.}
\label{fig:meanA}
\end{figure}

First, we use Algorithm \ref{algo:mean} to compute mean graphs of 50 observations associated with the letter 'A', 
at each of the three distortion levels. 
Fig. \ref{fig:meanA} shows the results. 
To match nodes across multiple graphs, one has to add several null nodes in the mean shape, as seen in the resulting means. 
The mean graphs resemble the letter 'A' in all three cases, despite a significant 
variability and distortions in the original data. 

\begin{figure}
\begin{center}
\includegraphics[width=1\linewidth]{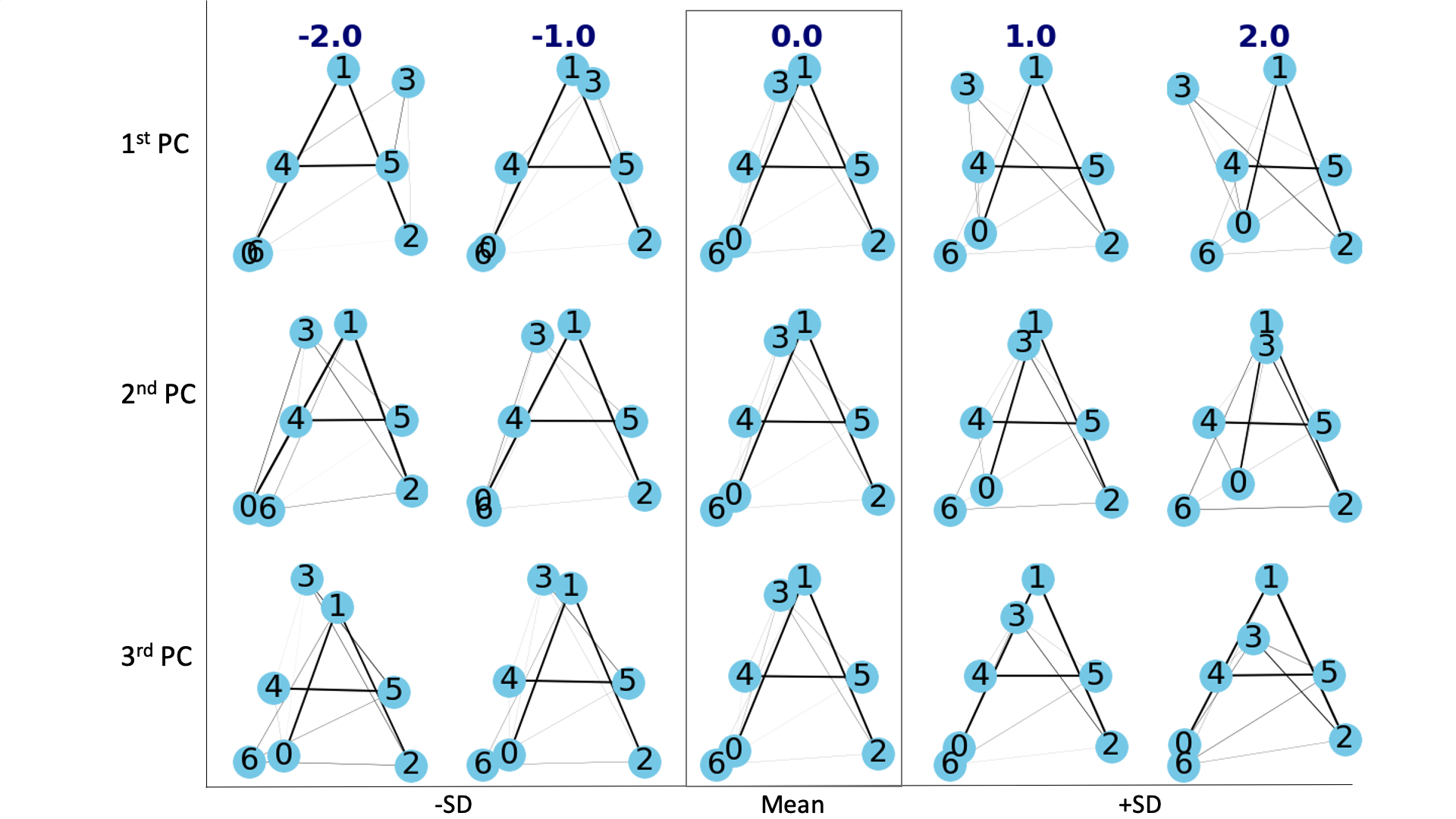} 
\end{center}
\caption{First three principal variations of letter 'A', low distortion. From top to bottom, each row shows the variation for the first, second and third principal directions, respectively. For each row, the middle one is mean while toward left and right, they are the graphs after perturbing the mean by one and two square root of singular value.}
\label{fig:pcaAlow}
\end{figure}

\begin{figure}
\begin{center}
\includegraphics[width=1\linewidth]{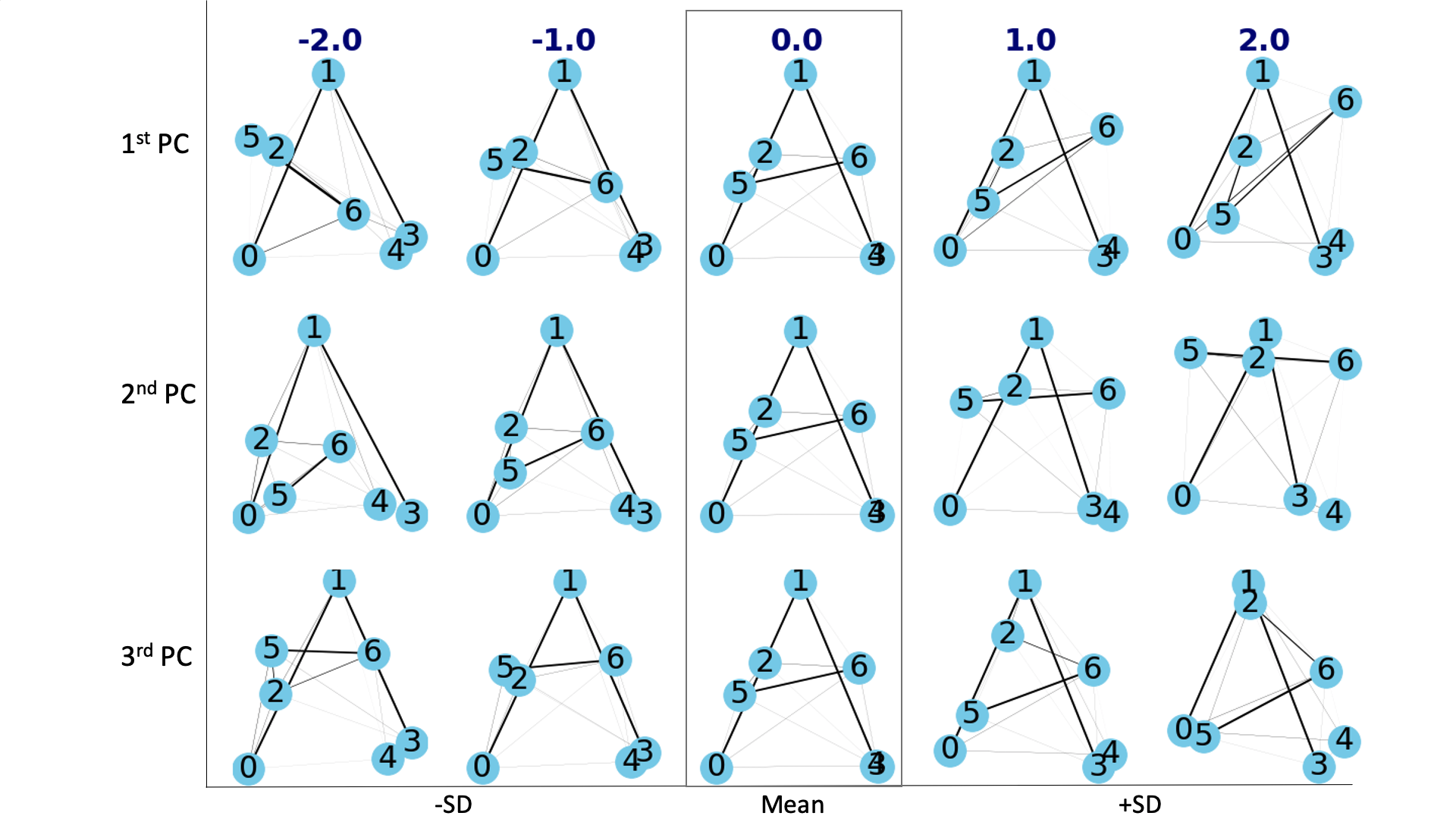}
\end{center}
\caption{Same as Fig. \ref{fig:pcaAlow}, medium distortion.}
\label{fig:pcaAmed}
\end{figure}

\begin{figure}
\begin{center}
\includegraphics[width=1\linewidth]{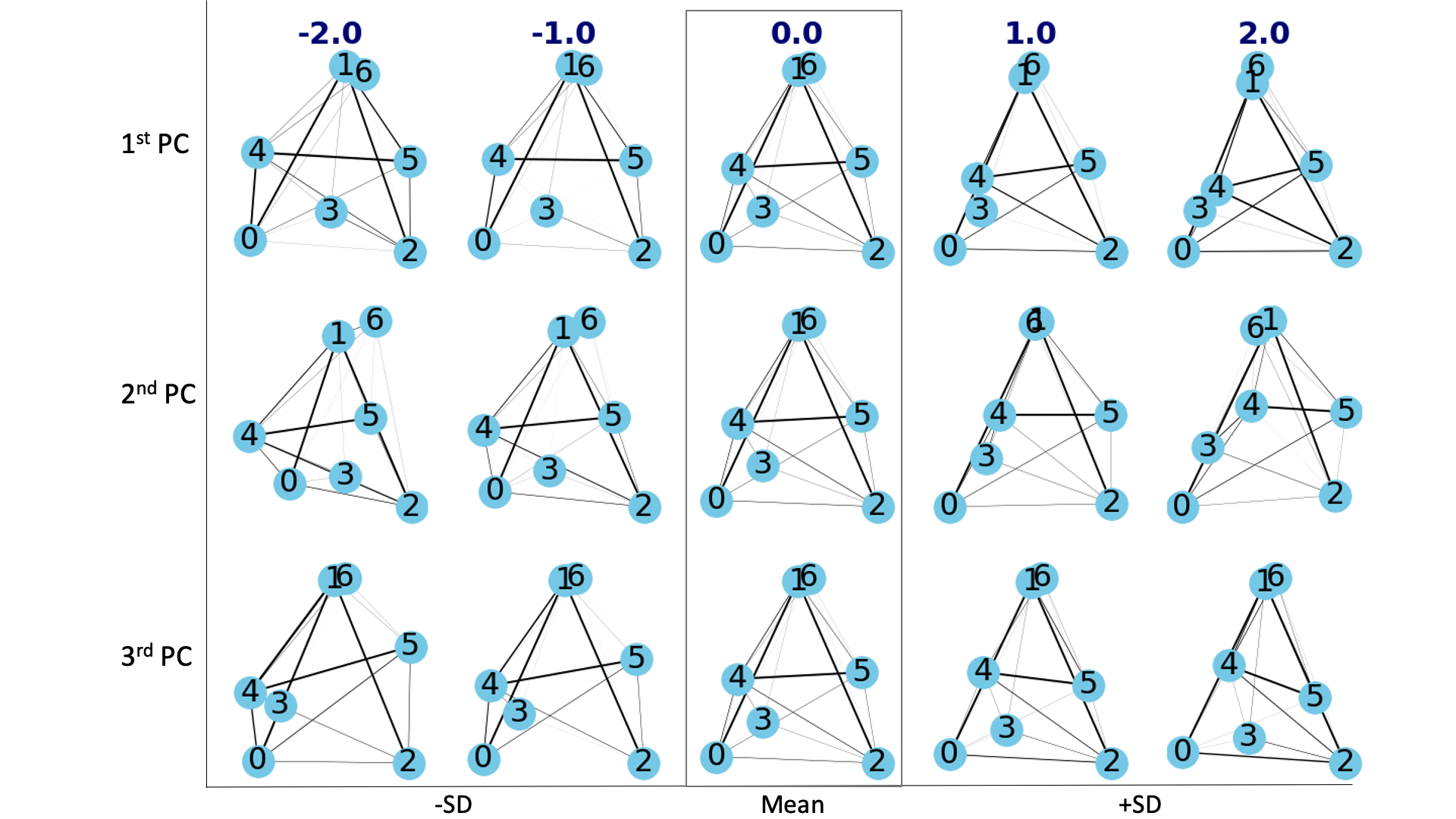}
\end{center}
\caption{Same as Fig. \ref{fig:pcaAlow}, high distortion.}
\label{fig:pcaAhigh}
\end{figure}

Additionally, we perform PCA on this letter data 
in the quotient space ${\cal G}$ and 
display results in 
Figs. \ref{fig:pcaAlow}, \ref{fig:pcaAmed} and \ref{fig:pcaAhigh}, for low, medium and high distortion
graphs, respectively. 
In these figures, each row depicts shape variability along a principal direction
in the given data in the form of graphs at mean $0, \pm 1, \pm 2$ standard deviation. 
This analysis helps identify the main modes of structural variability in the original data. 
For example, Fig. \ref{fig:pcaAlow} shows graphs along the first three principal directions of variability in the {\it low distortion} dataset. 
In all these graphs, the primary edges are stable, and there are no significant changes in principal directions. 
This implies that observations in this set are quite similar in shape.
However, the results for the 
medium distortion data shown in Fig. \ref{fig:pcaAmed} 
are different. The horizontal edge ${\it 5-6}$ changes significantly in the first principal direction.
In the high distortion level case, there are significant changes in shapes along all principal directions. For instance, there are extra edges in the top row (${\it 1-4}$ and ${\it 4-0}$) that dominate each principal direction. 

\begin{figure} 
\centering
\subfloat[Low Distortion]
{
\includegraphics[width=1\linewidth]{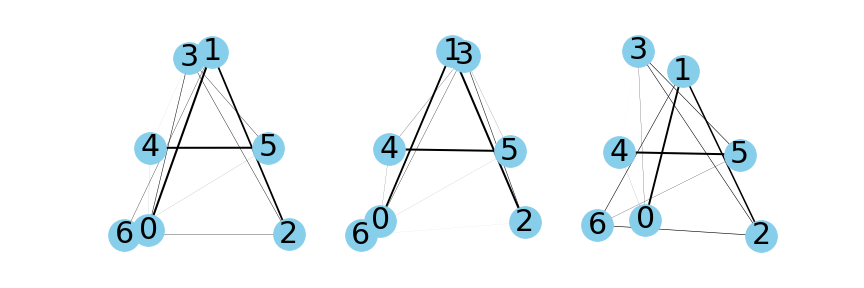}   
}\\
\subfloat[Medium Distortion]
{
\includegraphics[width=1\linewidth]{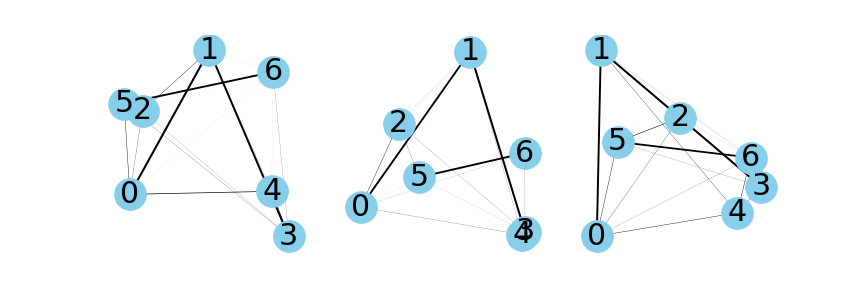}   
}\\
\subfloat[High Distortion]
{
\includegraphics[width=1\linewidth]{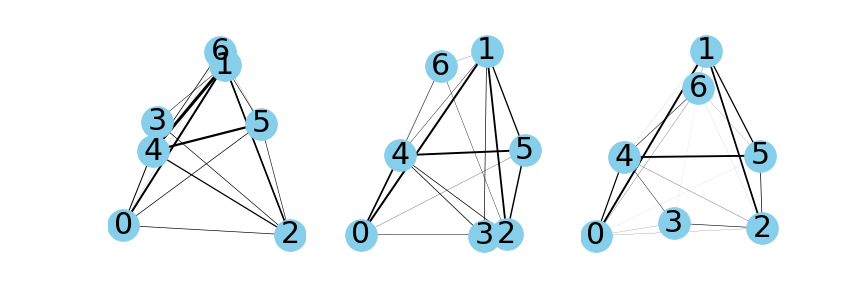}   
}
\caption{Random samples of letter 'A' from Gaussian models, for the three letter A datasets.}
\label{fig:RandomSampleA}
\end{figure}

Additionally, we fitted a Gaussian model on PC scores of these letter graphs. 
We first use PCA to reduce the dimension to capture approx. 80\% variance, resulting in
the first 6, 13, and 13 principal components out of 35 dimensions for low, medium, and high distortion letters, respectively.
Then, we impose a multivariate Gaussian model on these principal component scores. 
To evaluate this model, we generate some random samples from this model and project them into graph space, presented in Fig. \ref{fig:RandomSampleA}.
A visible similarity of these random samples to the original graphs underscores the goodness of the model. 

\begin{figure}
\begin{center}
\includegraphics[width=1\linewidth]{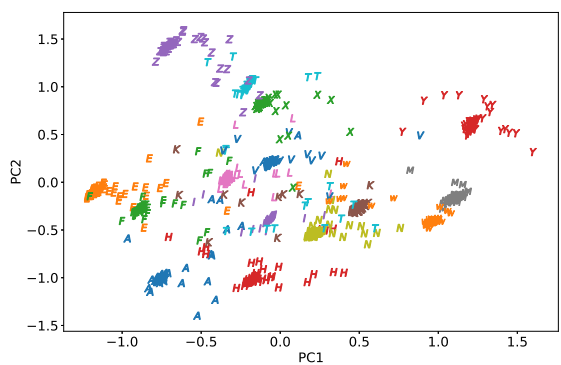}
\end{center}
\caption{Low distortion letter graphs project into two-dimensional principal subspace.}
\label{fig:letter_pca}
\end{figure}

Another vital application of PCA is dimension reduction -- perform PCA on all graph data and represent each one using its first few PC scores.
In this experiment, we project all 750 samples (50 samples in each letter class) into the first two principal scores and display them as their class symbols in Fig. \ref{fig:letter_pca}.
Although we do not use the class information in this PCA, 
note that low-dimensional representations of letter graphs separate well into different clusters, according to their classes. 

\subsection{Molecular Graphs}
\begin{figure}
\centering
\subfloat[Active]
{
\includegraphics[width=0.8\linewidth]{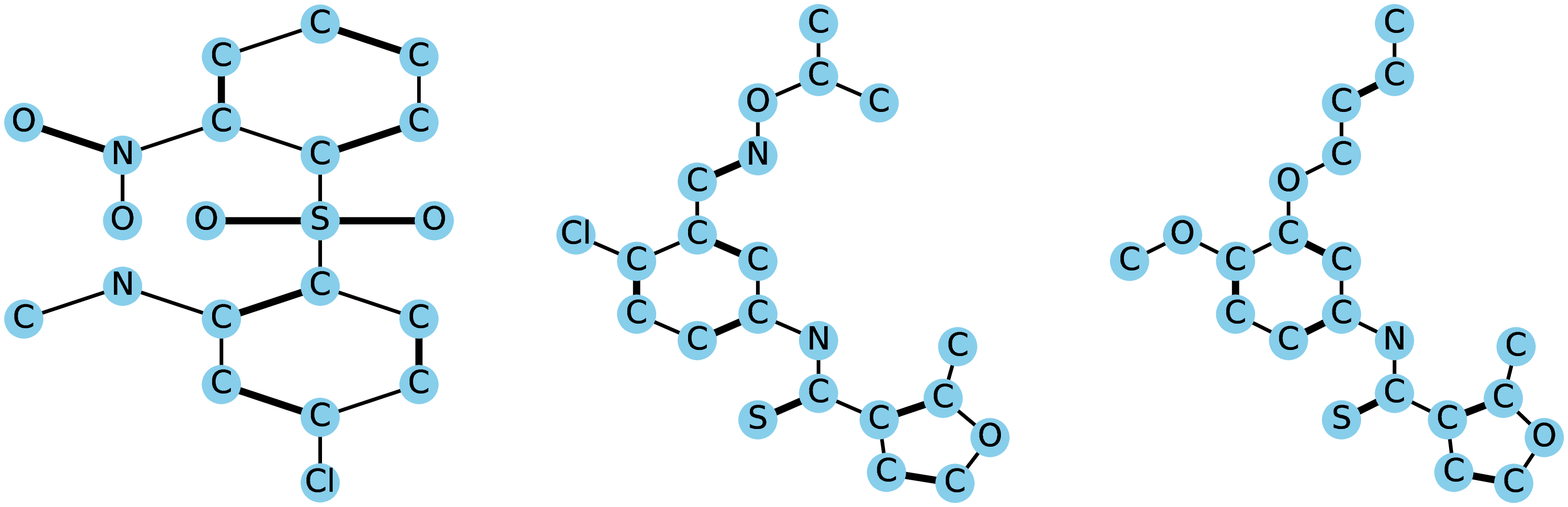}  
}\\
\subfloat[Inactive]
{
\includegraphics[width=0.8\linewidth]{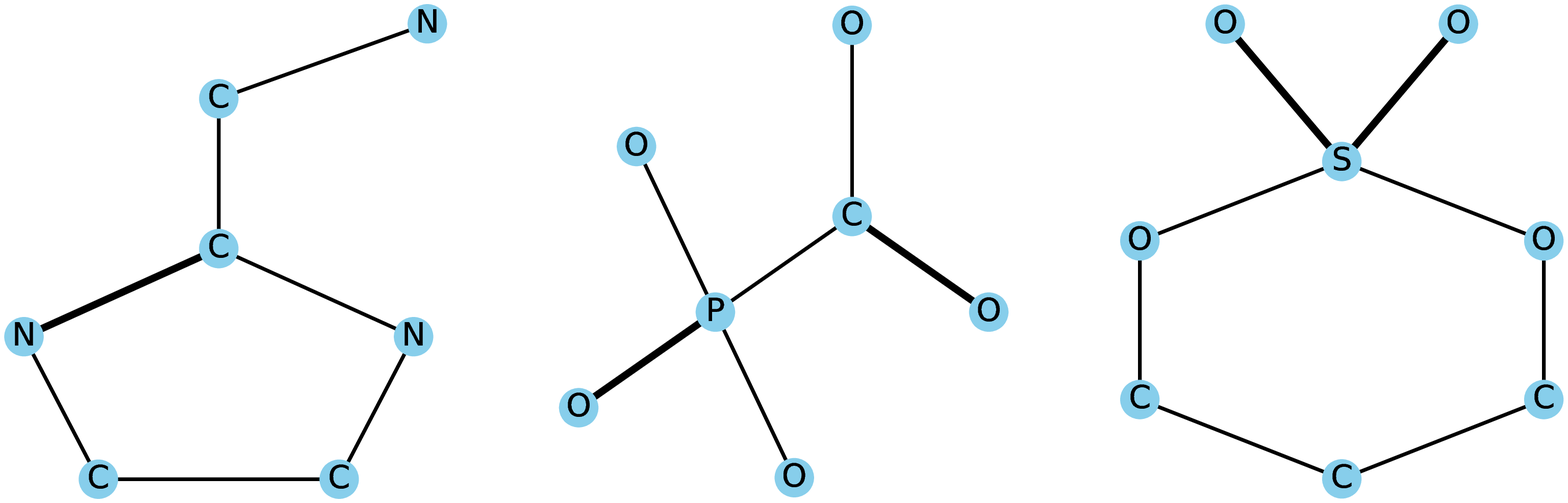}  
}
\caption{Sample graphs of molecules.}
\label{fig:molecule}
\end{figure}

\begin{figure}
\centering
\subfloat[Geodesic Interpolation in ${\cal A}$]
{
\begin{tabular}{c}
\hspace{-1cm}
\includegraphics[width=1.2\linewidth]{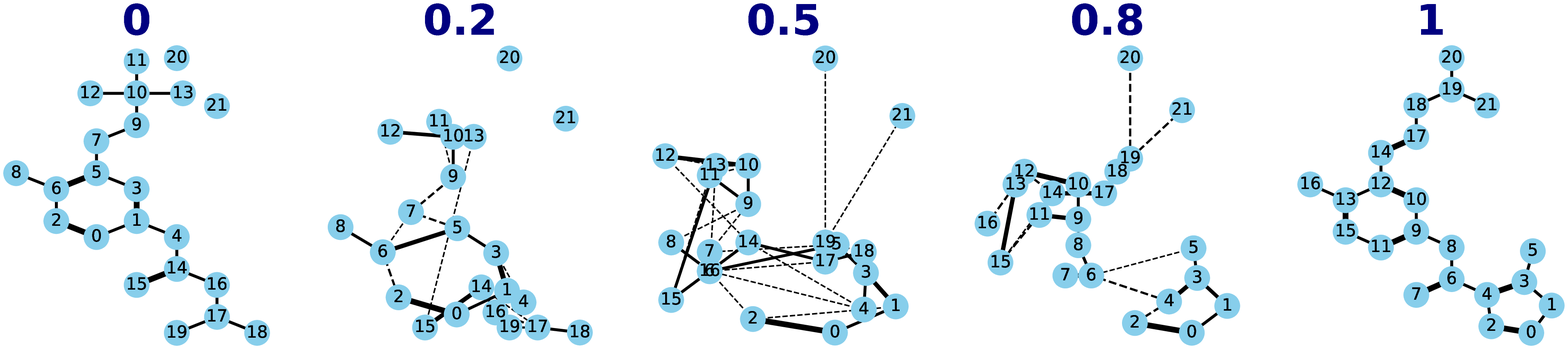}\\
\hspace{-1cm}
\includegraphics[width=1.2\linewidth]{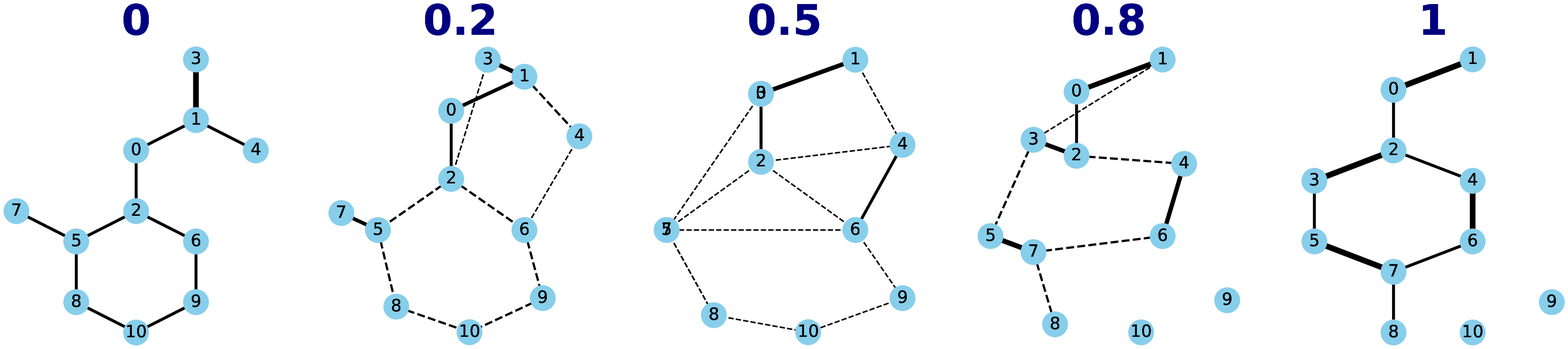}
\end{tabular}
}\\
\subfloat[Geodesic Interpolation in ${\cal G}$]
{
\begin{tabular}{c}
\hspace{-1cm}
\includegraphics[width=1.2\linewidth]{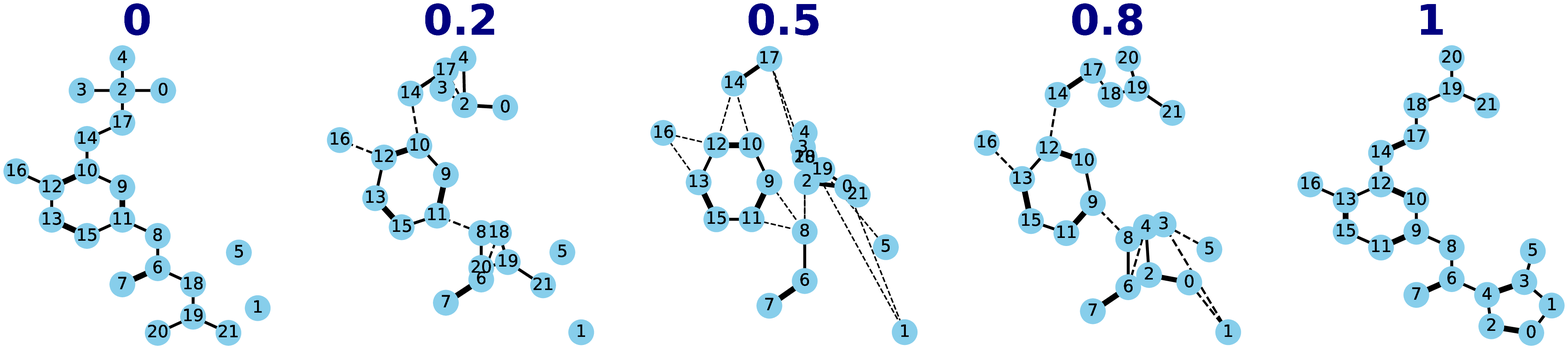}\\
\hspace{-1cm}
\includegraphics[width=1.2\linewidth]{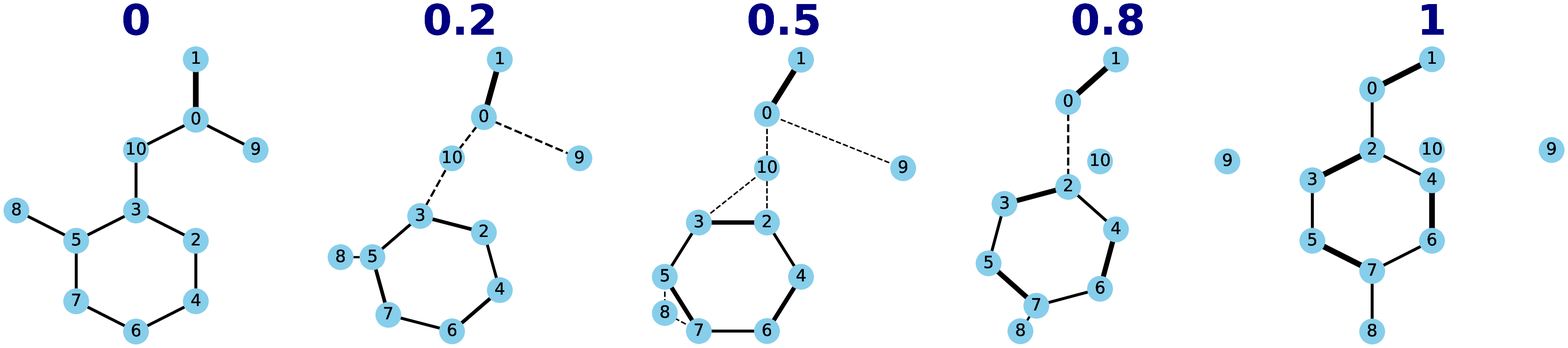}
\end{tabular}
}
\caption{Comparison between geodesics in original space and graph space for two different molecule graphs in the same class. In each subplot, the top is for active molecules, while the bottom is for inactive molecules. The time point is labeled on the top of each graph, while 0 and 1 indicate the original graphs. Dashed lines imply that the edges are changing. }
\label{fig:MoleculeGeodesic}
\end{figure}

In this section, we analyze another graph dataset from IAM Graph Database~\cite{riesen2008iam},  
this time involving molecular compounds. 
These molecules are straightforwardly converted into graphs by representing 
atoms as nodes and the chemical bonds as edges. 
This dataset consists of two classes (active, inactive), which denotes
whether the molecules are active against HIV or not. 
Fig. \ref{fig:molecule} shows some example graphs of active and inactive molecules.
We use the binary 
edge weights and atom labels (converted to one-hot vector) 
as node attributes in these experiments. 
Setting $\lambda =1$ for matching graphs, we present two pairs of geodesics in ${\cal A}$ and ${\cal G}$ in Fig. \ref{fig:MoleculeGeodesic}.
One can see that the path in ${\cal G}$ has a more natural deformation.

\begin{figure}
\centering
\subfloat[Active]
{
\includegraphics[width=0.45\linewidth]{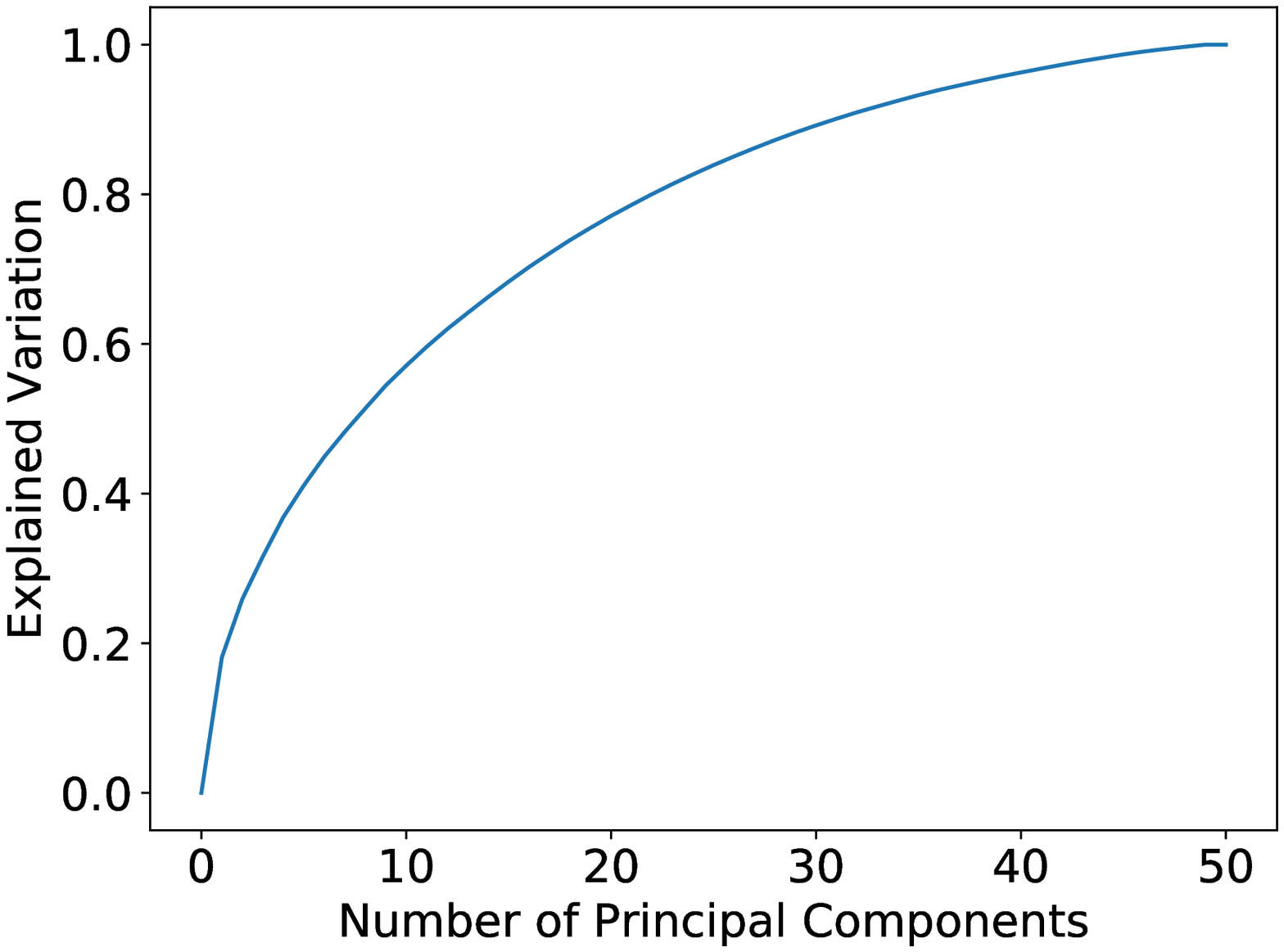}  
}
\subfloat[Inactive]
{
\includegraphics[width=0.45\linewidth]{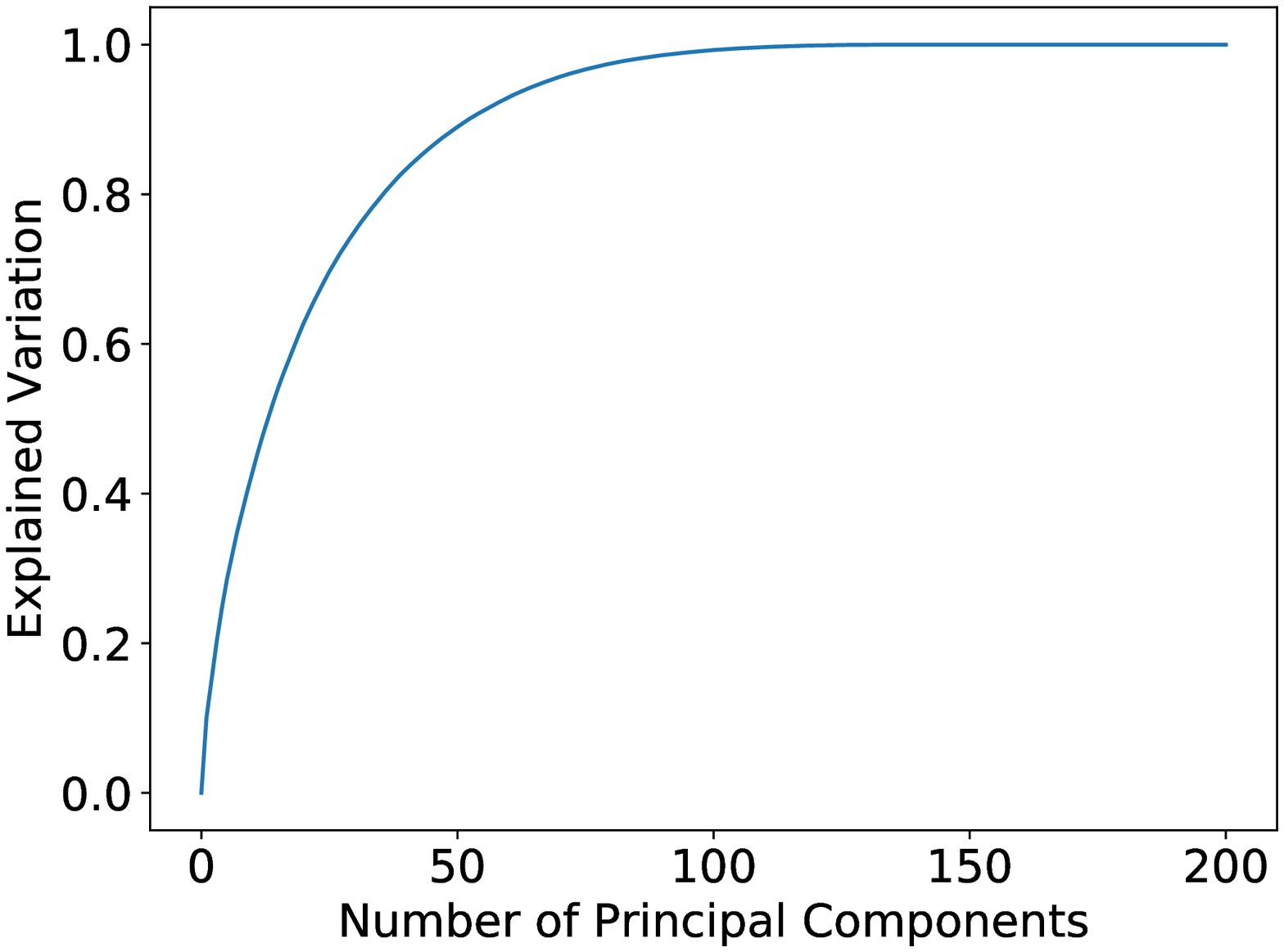}  
}
\caption{Cumulative explained variation of molecules by PCA. The vertical axis is the percentage of explained variation, while the horizontal axis is the number of principal components.}
\label{fig:pcaPercent_molecule}
\end{figure}

\begin{figure}
\centering
\includegraphics[width=1\linewidth]{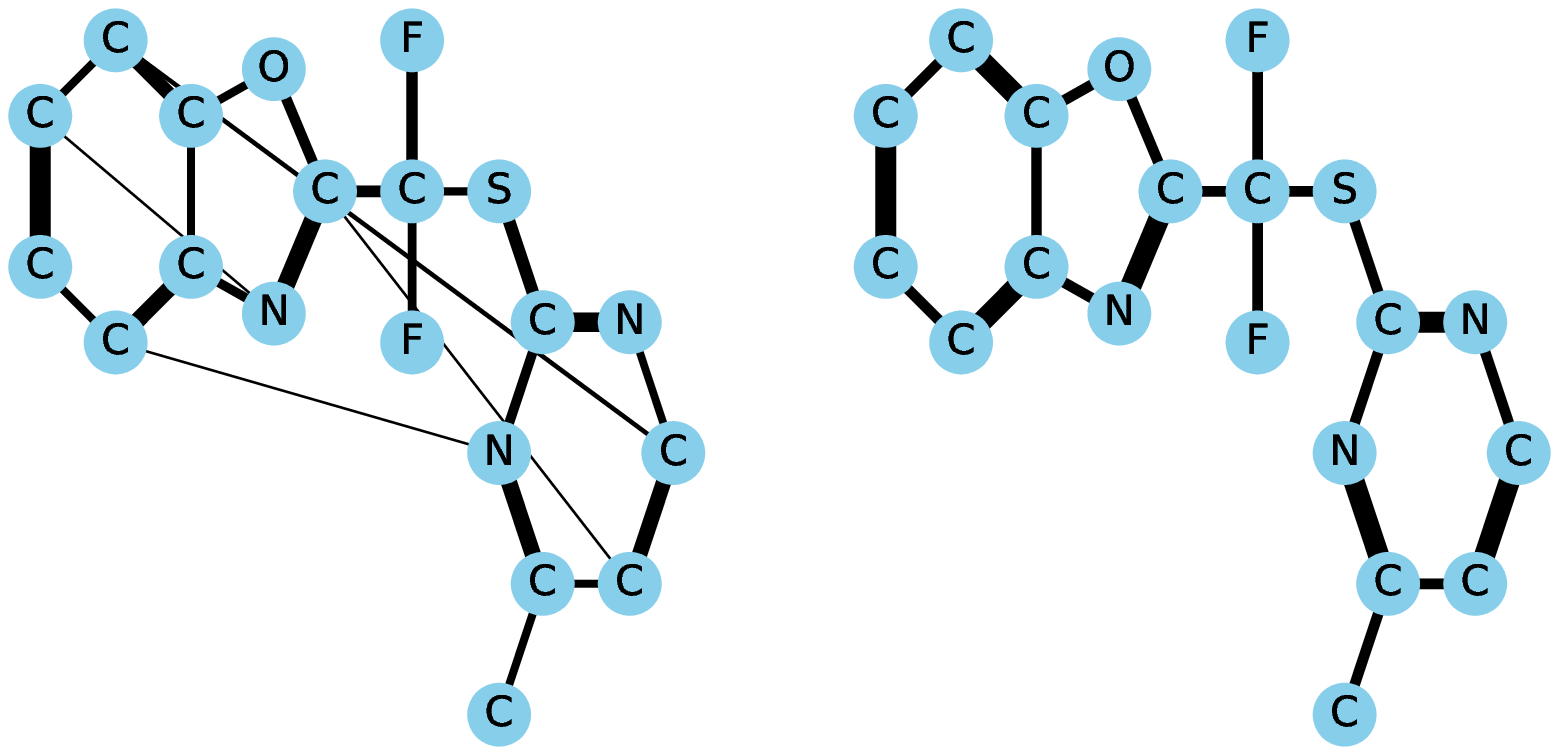}\\
\includegraphics[width=1\linewidth]{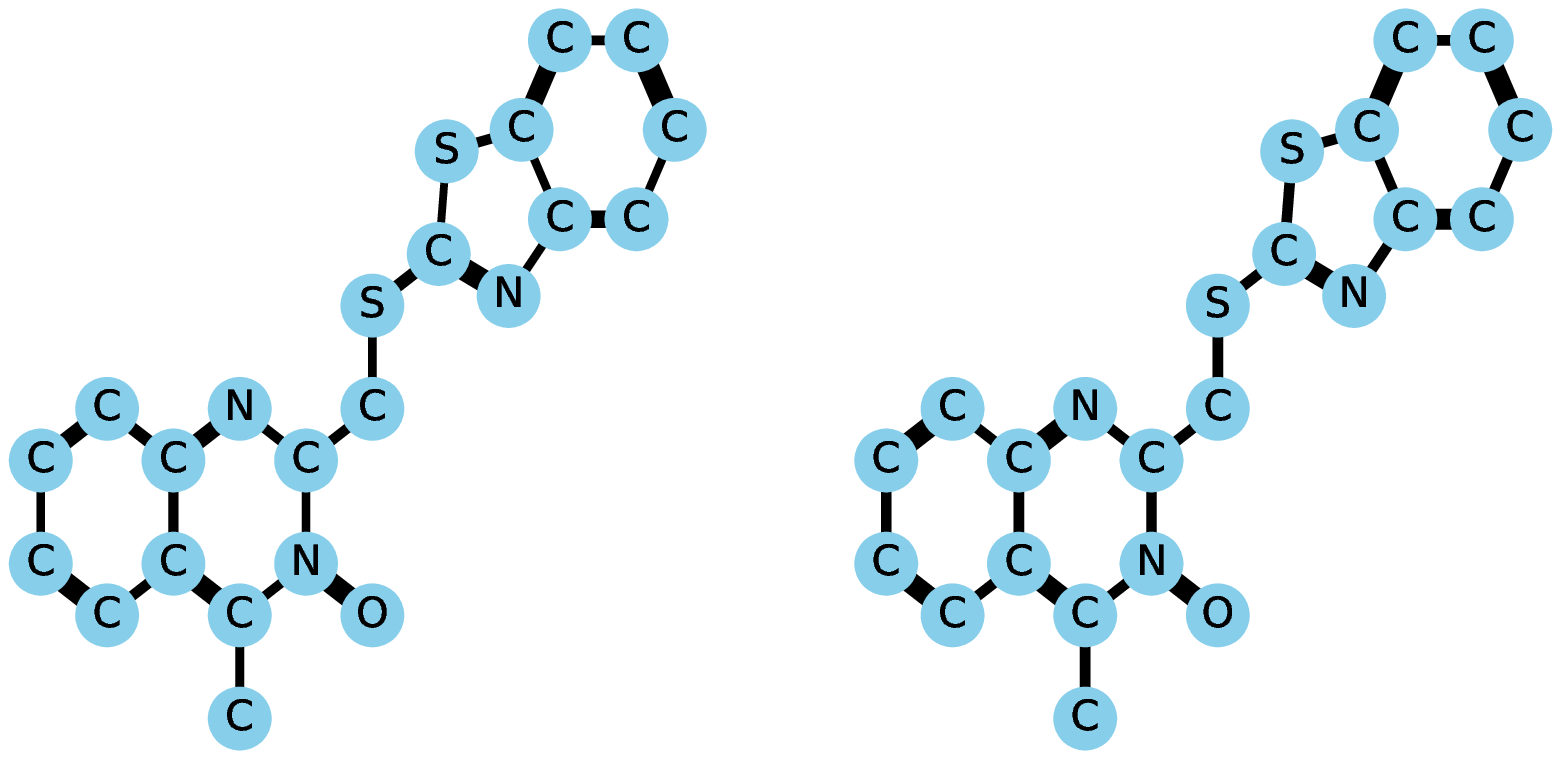}  
\caption{Reconstructed graphs of active molecules. For each pair, left is the reconstruction while right is the original graph. Some weak edges \textcolor{black}{(around 10\% edges weight)} have been removed to focus on the prominent edges.}
\label{fig:recon_mole}
\end{figure}

\begin{figure}
\centering
\includegraphics[width=1\linewidth]{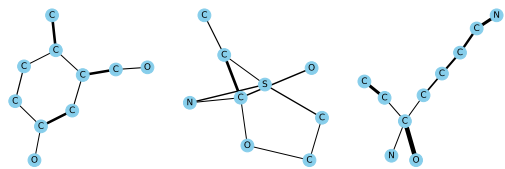}  
\caption{Random samples from the Gaussian model for inactive molecules.}
\label{fig:rs_mole}
\end{figure}

The complex structure of molecules results in the high-dimensional representation in the graph space ${\cal G}$, but we can reduce the representation size using graph PCA.
We perform a graph PCA on the molecule data as follows.
For the 50 active molecules, due to limited sample size, we only 
use edge weights (and not the node attributes) to perform graph PCA.
However, for 200 inactive modules, we utilize both edge weights and node attributes.
Fig. \ref{fig:pcaPercent_molecule} plots the captured variance versus the number of principal components. 
We use the first 22 and 36 principal components for active and inactive classes, 
respectively, containing roughly $80\%$ of the total variance to represent these molecules. 
For active classes, we reconstruct them from the principal scores. 
As shown in Fig. \ref{fig:recon_mole}, one can successfully reconstruct the original graphs with the chosen smaller dimensions. 
For inactive classes, we fit a multivariate Gaussian model to the principal scores. 
Fig. \ref{fig:rs_mole} shows some random samples generated from the model.
Since the model includes both node atoms and their chemical bonds, the sample
graphs contain both these attributes. 

\subsection{Wikipedia Graphs}
\begin{figure}[H]
\centering
\includegraphics[width=1.1\linewidth]{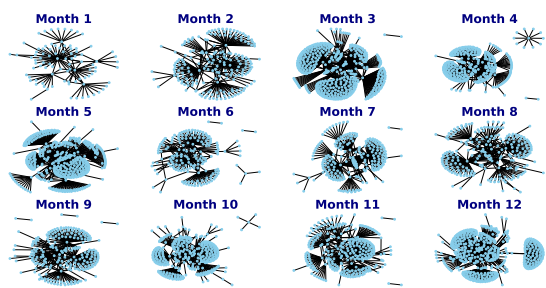} 
\caption{Wikipedia graphs}
\label{fig:wiki_sample}
\end{figure}

\begin{figure}[H]
\centering
\subfloat[Mean in ${\cal G}$]
{
\includegraphics[width=0.5\linewidth]{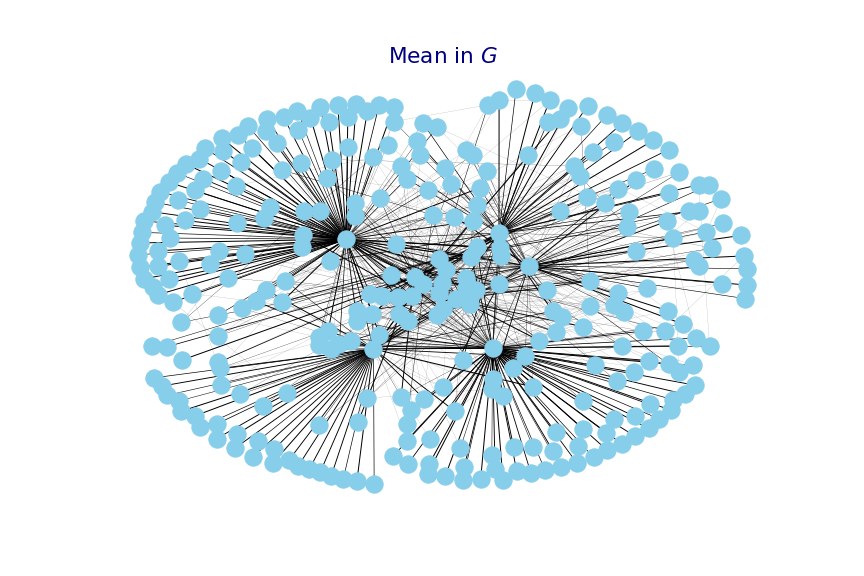}   
}
\subfloat[Energy vs Iteration]
{
\includegraphics[width=0.5\linewidth]{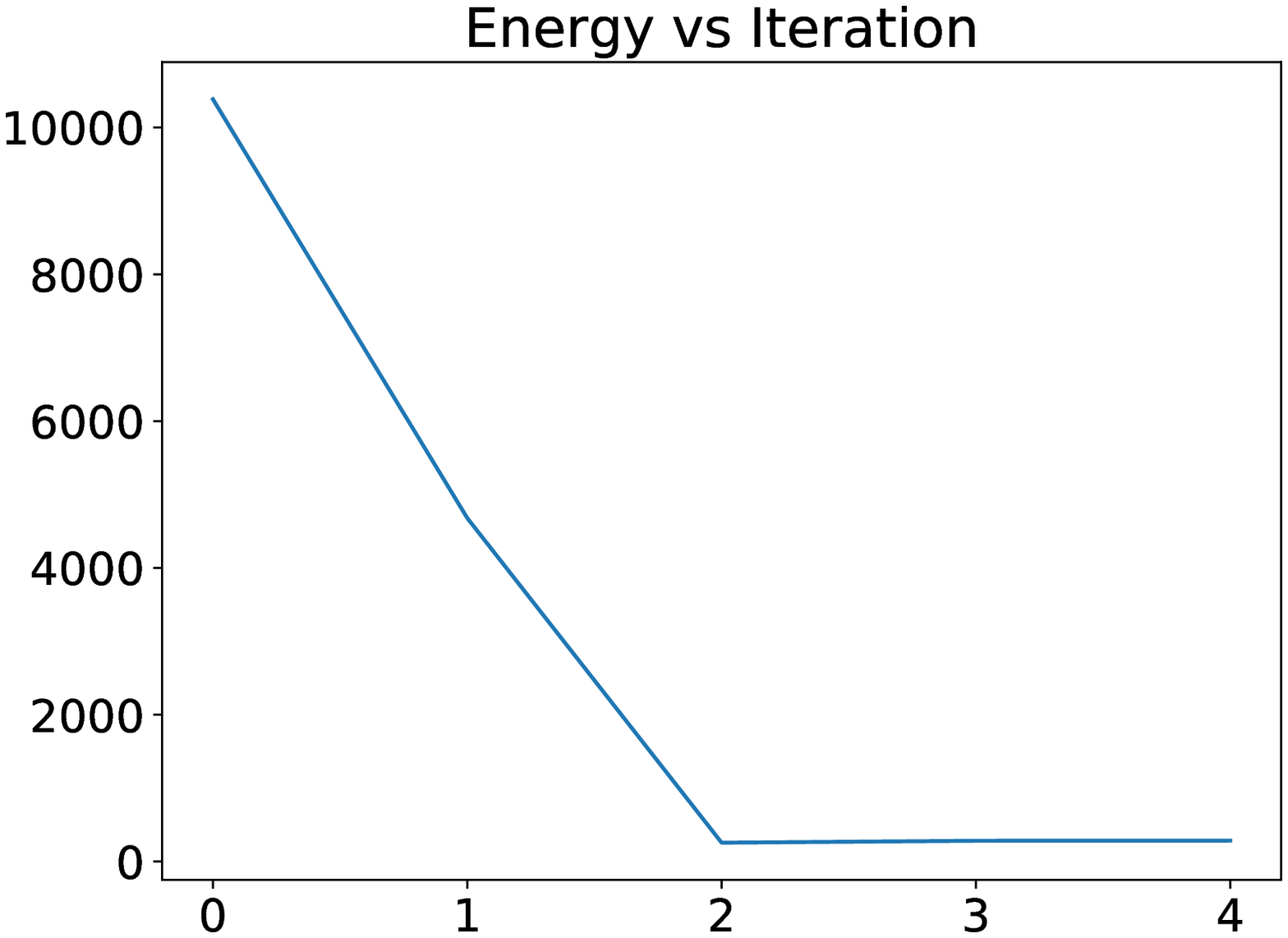}   
}

\caption{Karcher mean for Wikipedia graphs.}
\label{fig:wikiMean}
\end{figure}

Our last example comes from communication networks of the Chinese Wikipedia \cite{konect:2017:wiki_talk_zh, konect:sun_2016_49561}. 
In these graphs, the nodes represent the Chinese Wikipedia users, and an edge (0 or 1) denotes whether one user left a message on the talk page to another user at a certain timestamp. 
We emphasize that the goal here is to study connectivity patterns between users and the actual user identity is not important.
We take monthly graphs from the year 2004, resulting in a sample size of 12 graphs, see Fig \ref{fig:wiki_sample}.
On average, each graph has around 300 nodes and 431 edges. 
We compute the mean in graph space ${\cal G}$, with results shown in Fig. \ref{fig:wikiMean}.
This graph 
shows a clear clustering of users, implying that major subsets of users actively interact with others in their clusters.

\begin{figure*}
\begin{center}
\includegraphics[width=1\linewidth]{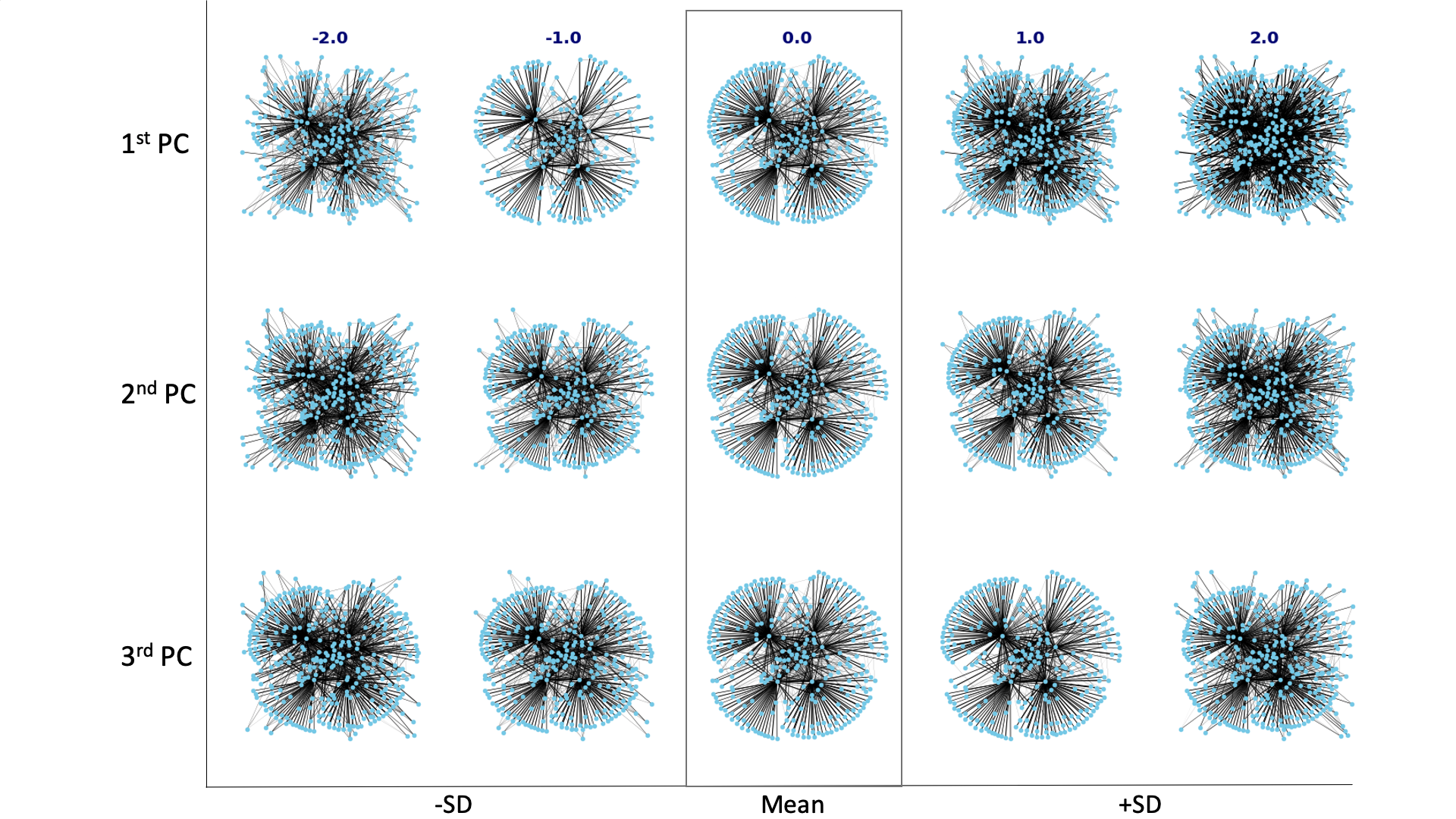}
\end{center}
\caption{First three principal variations of Wikipedia graphs. From top to bottom, each row shows the variation for the first, second and third principal directions, respectively. The middle graph in each row is the mean, while left and right graphs result from perturbing the mean by one and two square root of singular value.}
\label{fig:wikiPCA}
\end{figure*}

The results from PCA analysis of these graphs are shown in Fig. \ref{fig:wikiPCA}. These results show that most of the user interactions are stable and 
remain unchanged, while principal variations in the data come only from a handful of active users.  

\subsection{Classification}
Even though our prime goal is to develop analytical generative models for graph data, the framework also easily applies to classification problems.  
Here we present classification results on some benchmark graph datasets using the proposed 
framework as follows. 
We use SVM with Radial Basis Function (RBF) kernel as the classifier and distance vectors 
$d_g$ (between an unlabelled test graph and a set of labeled training graphs) as the input. 
We first report the results on the letter and molecule graphs from IAM database~\cite{riesen2008iam} in Table \ref{tab:letter_mole}.
The data have already been split into training, validation, and test set by the owners.
So we report results on the same test set as in~\cite{riesen2008iam}.
Additionally, we also apply this method on some graph datasets from the popular TU database~\cite{KKMMN2016}. 
For this, we adopt the nested five-fold cross-validation technique, similar to \cite{kriege2020survey, vayer2018optimal}. 
The hyperparameters of SVM are searched on a grid in the inner loop, 
while the results are reported on the test folds of the outer loop.
We compare the results with \cite{vayer2018optimal} that uses optimal transport-based distances and the interquartile interval of corresponding results from different graph kernels in \cite{kriege2020survey}.
(One can find a larger comparison of graph neural networks in \cite{errica2019fair}.)
As the results show, the proposed metric achieves classification results that are comparable with other published methods. 

\begin{table*}
\caption{Letter and molecule graph classification results}
\label{tab:letter_mole}
\begin{center}
\begin{tabular}{c|c|c|c|c}
\hline
  & {\scriptsize Letter: Low Distortion} &{\scriptsize Letter: Medium Distortion} & {\scriptsize Letter: High Distortion} & Molecule \\
\hline\hline
Ours & 98.5\% & 96.4\% & 93.6\% & 99.6\%\\
\cite{riesen2008iam}  & 99.6\% & 94.0\% & 90.0\% & 97.3\%\\
\hline
\end{tabular}
\end{center}
\end{table*}

\begin{table*}
\caption{Classification results on some TU graph datasets}
\label{tab:classification}
\begin{center}
\begin{tabular}{c|c|c|c|c|c|c}
\hline
\% & BZR & IMDB-B & IMDB-M & MUTAG & PROTEIN &PTC-MR\\
 \hline
 \hline 
Ours &83.21$\pm$2.54 & 71.60$\pm$3.86 & 49.73$\pm$3.49 & 83.50$\pm$7.57 & 74.84 $\pm$ 3.05
& 59.84$\pm$7.34 \\
\cite{vayer2018optimal} & 85.12$\pm$4.15 & 63.80$\pm$3.49 & 48.00$\pm$3.22 & 83.26$\pm$10.30 & 74.55$\pm$2.74 & 55.71$\pm$6.74 \\
(Quartile 1, Quartile 3)\cite{kriege2020survey} & (85.2, 88.0) & (61.0, 71.2) &(42.1, 49.3) & (83.7, 87.3)&(74.1, 75.7)&(59.0, 61.0) \\
\hline
\end{tabular}
\end{center}
\end{table*}

\section{Discussion \& Conclusion}
\label{sec:conclusion}
In this paper, we 
build on the foundation laid by Jain et al. and 
develop a statistical framework for learning and analyzing structures of graphs. 
The quotient space formulation removes the nuance permutation variability and helps register 
nodes across graphs in a natural way.
Due to the isometric action of the permutation group, 
the quotient space inherits metric that enables metric-based statistical 
analysis of graphs -- geodesics, means, PCA, and Gaussian-type models. 

The set of tools developed in this paper are useful in several contexts. For instance, one can use them to analyze geometrical deep learning methods, 
where both data and inferences can involve graphs in different forms. Low-dimensional Euclidean representations of graphs will enable direct use of more sophisticated statistical models, including many deep learning architectures. The ability to reconstruct full graphs from these representations is vital in synthesizing new graphs. 


\begin{acknowledgements}
The authors would like to thank the creators of different public datasets used in this paper. The authors also thank Dr. Adam Duncan for his contributions
in the implementation of a preliminary version of the approach and Dr. Derek Tucker for his contribution in the Python implementation of FAQ algorithm. 
This research was supported in part by the grants NSF CDS\&E DMS 1953087
and NSF IIS 1955154 to AS, and NSF IIS 1956050 to SS. 
\end{acknowledgements}

%
%

\bibliographystyle{spbasic}   
\bibliography{bib.bib}  

\begin{thebibliography}{68}
\providecommand{\natexlab}[1]{#1}
\providecommand{\url}[1]{{#1}}
\providecommand{\urlprefix}{URL }
\expandafter\ifx\csname urlstyle\endcsname\relax
  \providecommand{\doi}[1]{DOI~\discretionary{}{}{}#1}\else
  \providecommand{\doi}{DOI~\discretionary{}{}{}\begingroup
  \urlstyle{rm}\Url}\fi
\providecommand{\eprint}[2][]{\url{#2}}

\bibitem[{kon(2017)}]{konect:2017:wiki_talk_zh}
 (2017) Wikipedia talk, chinese network dataset -- {KONECT}.
  \urlprefix\url{http://konect.uni-koblenz.de/networks/wiki_talk_zh}

\bibitem[{Almohamad and Duffuaa(1993)}]{almohamad1993linear}
Almohamad H, Duffuaa SO (1993) A linear programming approach for the weighted
  graph matching problem. IEEE Transactions on pattern analysis and machine
  intelligence 15(5):522--525

\bibitem[{Bahonar et~al.(2019)Bahonar, Mirzaei, Sadri, and
  Wilson}]{bahonar2019graph}
Bahonar H, Mirzaei A, Sadri S, Wilson R (2019) Graph embedding using frequency
  filtering. IEEE transactions on pattern analysis and machine intelligence

\bibitem[{Bai et~al.(2019)Bai, Ding, Qiao, Marinovic, Gu, Chen, Sun, and
  Wang}]{baiunsupervised}
Bai Y, Ding H, Qiao Y, Marinovic A, Gu K, Chen T, Sun Y, Wang W (2019)
  Unsupervised inductive graph-level representation learning via graph-graph
  proximity. In: IJCAI

\bibitem[{Caelli and Kosinov(2004)}]{caelli2004eigenspace}
Caelli T, Kosinov S (2004) An eigenspace projection clustering method for
  inexact graph matching. IEEE transactions on pattern analysis and machine
  intelligence 26(4):515--519

\bibitem[{Carletti et~al.(2017)Carletti, Foggia, Saggese, and
  Vento}]{carletti2017challenging}
Carletti V, Foggia P, Saggese A, Vento M (2017) Challenging the time complexity
  of exact subgraph isomorphism for huge and dense graphs with vf3. IEEE
  transactions on pattern analysis and machine intelligence 40(4):804--818

\bibitem[{Chapel et~al.(2020)Chapel, Alaya, and Gasso}]{chapel2020partial}
Chapel L, Alaya MZ, Gasso G (2020) Partial optimal transport with applications
  on positive-unlabeled learning. arXiv preprint arXiv:200208276

\bibitem[{Chen et~al.(2018)Chen, Perozzi, Hu, and Skiena}]{chen2018harp}
Chen H, Perozzi B, Hu Y, Skiena S (2018) Harp: Hierarchical representation
  learning for networks. In: Thirty-Second AAAI Conference on Artificial
  Intelligence

\bibitem[{Chizat et~al.(2018)Chizat, Peyr{\'e}, Schmitzer, and
  Vialard}]{chizat2018scaling}
Chizat L, Peyr{\'e} G, Schmitzer B, Vialard FX (2018) Scaling algorithms for
  unbalanced optimal transport problems. Mathematics of Computation
  87(314):2563--2609

\bibitem[{Chowdhury and Needham(2019)}]{chowdhury2019gromov}
Chowdhury S, Needham T (2019) Gromov-wasserstein averaging in a riemannian
  framework. arXiv preprint arXiv:191004308

\bibitem[{Conte et~al.(2004)Conte, Foggia, Sansone, and
  Vento}]{conte2004thirty}
Conte D, Foggia P, Sansone C, Vento M (2004) Thirty years of graph matching in
  pattern recognition. International journal of pattern recognition and
  artificial intelligence 18(03):265--298

\bibitem[{Dai et~al.(2016)Dai, Zhang, and Srivastava}]{dai2016testing}
Dai M, Zhang Z, Srivastava A (2016) Testing stationarity of brain functional
  connectivity using change-point detection in fmri data. In: Proceedings of
  the IEEE Conference on Computer Vision and Pattern Recognition Workshops, pp
  19--27

\bibitem[{De~Souza et~al.(2014)De~Souza, Sarkar, Srivastava, and
  Su}]{de2014pattern}
De~Souza FD, Sarkar S, Srivastava A, Su J (2014) Pattern theory-based
  interpretation of activities. In: Pattern Recognition (ICPR), 2014 22nd
  International Conference on, IEEE, pp 106--111

\bibitem[{Defferrard et~al.(2016)Defferrard, Bresson, and
  Vandergheynst}]{defferrard2016convolutional}
Defferrard M, Bresson X, Vandergheynst P (2016) Convolutional neural networks
  on graphs with fast localized spectral filtering. In: Advances in neural
  information processing systems, pp 3844--3852

\bibitem[{Dryden and Mardia(1998)}]{mardia-dryden-book}
Dryden IL, Mardia KV (1998) Statistical Shape Analysis. John Wiley \& Son

\bibitem[{Errica et~al.(2019)Errica, Podda, Bacciu, and
  Micheli}]{errica2019fair}
Errica F, Podda M, Bacciu D, Micheli A (2019) A fair comparison of graph neural
  networks for graph classification. arXiv preprint arXiv:191209893

\bibitem[{Frank and Wolfe(1956)}]{frank1956algorithm}
Frank M, Wolfe P (1956) An algorithm for quadratic programming. Naval research
  logistics quarterly 3(1-2):95--110

\bibitem[{Gold and Rangarajan(1996)}]{gold1996graduated}
Gold S, Rangarajan A (1996) A graduated assignment algorithm for graph
  matching. IEEE Transactions on pattern analysis and machine intelligence
  18(4):377--388

\bibitem[{Gramfort et~al.(2015)Gramfort, Peyr{\'e}, and
  Cuturi}]{gramfort2015fast}
Gramfort A, Peyr{\'e} G, Cuturi M (2015) Fast optimal transport averaging of
  neuroimaging data. In: International Conference on Information Processing in
  Medical Imaging, Springer, pp 261--272

\bibitem[{Grover and Leskovec(2016)}]{grover2016node2vec}
Grover A, Leskovec J (2016) node2vec: Scalable feature learning for networks.
  In: Proceedings of the 22nd ACM SIGKDD international conference on Knowledge
  discovery and data mining, ACM, pp 855--864

\bibitem[{Guo et~al.(2020)Guo, Bal, Needham, and Srivastava}]{guo-Brain:2020}
Guo X, Bal AB, Needham T, Srivastava A (2020) Statistical shape analysis of
  brain arterial networks (bans). arXiv:200704793

\bibitem[{Hakimi(1965)}]{hakimi1965optimum}
Hakimi SL (1965) Optimum distribution of switching centers in a communication
  network and some related graph theoretic problems. Operations research
  13(3):462--475

\bibitem[{Hamilton et~al.(2017)Hamilton, Ying, and
  Leskovec}]{hamilton2017inductive}
Hamilton W, Ying Z, Leskovec J (2017) Inductive representation learning on
  large graphs. In: Advances in Neural Information Processing Systems, pp
  1024--1034

\bibitem[{Han et~al.(2015)Han, Wilson, and Hancock}]{han2015generative}
Han L, Wilson RC, Hancock ER (2015) Generative graph prototypes from
  information theory. IEEE transactions on pattern analysis and machine
  intelligence 37(10):2013--2027

\bibitem[{Horaud(2012)}]{horaud2012short}
Horaud R (2012) A short tutorial on graph laplacians, laplacian embedding, and
  spectral clustering

\bibitem[{Jain(2016{\natexlab{a}})}]{jain2016geometry}
Jain BJ (2016{\natexlab{a}}) On the geometry of graph spaces. Discrete Applied
  Mathematics 214:126--144

\bibitem[{Jain(2016{\natexlab{b}})}]{jain2016statistical}
Jain BJ (2016{\natexlab{b}}) Statistical graph space analysis. Pattern
  Recognition 60:802--812

\bibitem[{Jain and Obermayer(2009)}]{jain2009structure}
Jain BJ, Obermayer K (2009) Structure spaces. Journal of Machine Learning
  Research 10(Nov):2667--2714

\bibitem[{Jain and Obermayer(2012)}]{jain2012learning}
Jain BJ, Obermayer K (2012) Learning in riemannian orbifolds. arXiv preprint
  arXiv:12044294

\bibitem[{Kersting et~al.(2020)Kersting, Kriege, Morris, Mutzel, and
  Neumann}]{KKMMN2016}
Kersting K, Kriege NM, Morris C, Mutzel P, Neumann M (2020) Benchmark data sets
  for graph kernels. \urlprefix\url{http://www.graphlearning.io/}

\bibitem[{Kipf and Welling(2016)}]{kipf2016semi}
Kipf TN, Welling M (2016) Semi-supervised classification with graph
  convolutional networks. arXiv preprint arXiv:160902907

\bibitem[{Kolaczyk et~al.(2020)Kolaczyk, Lin, Rosenberg, Walters, Xu
  et~al.}]{kolaczyk2020averages}
Kolaczyk ED, Lin L, Rosenberg S, Walters J, Xu J, et~al. (2020) Averages of
  unlabeled networks: Geometric characterization and asymptotic behavior. The
  Annals of Statistics 48(1):514--538

\bibitem[{Krcmar and Dhawan(1994)}]{krcmar1994application}
Krcmar M, Dhawan A (1994) Application of genetic algorithms in graph matching.
  In: Proceedings of 1994 IEEE International Conference on Neural Networks
  (ICNN'94), IEEE, vol~6, pp 3872--3876

\bibitem[{Kriege et~al.(2020)Kriege, Johansson, and Morris}]{kriege2020survey}
Kriege NM, Johansson FD, Morris C (2020) A survey on graph kernels. Applied
  Network Science 5(1):1--42

\bibitem[{Kuhn(1955)}]{kuhn1955hungarian}
Kuhn HW (1955) The hungarian method for the assignment problem. Naval research
  logistics quarterly 2(1-2):83--97

\bibitem[{Lyzinski et~al.(2016)Lyzinski, Fishkind, Fiori, Vogelstein, Priebe,
  and Sapiro}]{lyzinski2016graph}
Lyzinski V, Fishkind DE, Fiori M, Vogelstein JT, Priebe CE, Sapiro G (2016)
  Graph matching: Relax at your own risk. IEEE transactions on pattern analysis
  and machine intelligence 38(1):60--73

\bibitem[{Mackaness and Beard(1993)}]{mackaness1993use}
Mackaness WA, Beard KM (1993) Use of graph theory to support map
  generalization. Cartography and Geographic Information Systems 20(4):210--221

\bibitem[{Maron and Lipman(2018)}]{maron2018probably}
Maron H, Lipman Y (2018) (probably) concave graph matching. arXiv preprint
  arXiv:180709722

\bibitem[{M{\'e}moli(2011)}]{memoli2011gromov}
M{\'e}moli F (2011) Gromov--wasserstein distances and the metric approach to
  object matching. Foundations of computational mathematics 11(4):417--487

\bibitem[{Narayanan et~al.(2017)Narayanan, Chandramohan, Venkatesan, Chen, Liu,
  and Jaiswal}]{narayanan2017graph2vec}
Narayanan A, Chandramohan M, Venkatesan R, Chen L, Liu Y, Jaiswal S (2017)
  graph2vec: Learning distributed representations of graphs. arXiv preprint
  arXiv:170705005

\bibitem[{Neuhaus and Bunke(2007)}]{neuhaus2007bridging}
Neuhaus M, Bunke H (2007) Bridging the gap between graph edit distance and
  kernel machines, vol~68. World Scientific

\bibitem[{Ortega et~al.(2018)Ortega, Frossard, Kova{\v{c}}evi{\'c}, Moura, and
  Vandergheynst}]{ortega2018graph}
Ortega A, Frossard P, Kova{\v{c}}evi{\'c} J, Moura JM, Vandergheynst P (2018)
  Graph signal processing: Overview, challenges, and applications. Proceedings
  of the IEEE 106(5):808--828

\bibitem[{Pele and Werman(2008)}]{pele2008linear}
Pele O, Werman M (2008) A linear time histogram metric for improved sift
  matching. In: European conference on computer vision, Springer, pp 495--508

\bibitem[{Perozzi et~al.(2014)Perozzi, Al-Rfou, and
  Skiena}]{perozzi2014deepwalk}
Perozzi B, Al-Rfou R, Skiena S (2014) Deepwalk: Online learning of social
  representations. In: Proceedings of the 20th ACM SIGKDD international
  conference on Knowledge discovery and data mining, ACM, pp 701--710

\bibitem[{Peyr{\'e} et~al.(2016)Peyr{\'e}, Cuturi, and
  Solomon}]{peyre2016gromov}
Peyr{\'e} G, Cuturi M, Solomon J (2016) Gromov-wasserstein averaging of kernel
  and distance matrices. In: International Conference on Machine Learning, pp
  2664--2672

\bibitem[{Riesen and Bunke(2008)}]{riesen2008iam}
Riesen K, Bunke H (2008) Iam graph database repository for graph based pattern
  recognition and machine learning. In: Joint IAPR International Workshops on
  Statistical Techniques in Pattern Recognition (SPR) and Structural and
  Syntactic Pattern Recognition (SSPR), Springer, pp 287--297

\bibitem[{Riesen and Bunke(2009)}]{riesen2009approximate}
Riesen K, Bunke H (2009) Approximate graph edit distance computation by means
  of bipartite graph matching. Image and Vision computing 27(7):950--959

\bibitem[{Sato(2020)}]{sato2020survey}
Sato R (2020) A survey on the expressive power of graph neural networks. arXiv
  preprint arXiv:200304078

\bibitem[{Sch{\"o}lkopf et~al.(1998)Sch{\"o}lkopf, Smola, and
  M{\"u}ller}]{scholkopf1998nonlinear}
Sch{\"o}lkopf B, Smola A, M{\"u}ller KR (1998) Nonlinear component analysis as
  a kernel eigenvalue problem. Neural computation 10(5):1299--1319

\bibitem[{S{\'e}journ{\'e} et~al.(2020)S{\'e}journ{\'e}, Vialard, and
  Peyr{\'e}}]{sejourne2020unbalanced}
S{\'e}journ{\'e} T, Vialard FX, Peyr{\'e} G (2020) The unbalanced gromov
  wasserstein distance: Conic formulation and relaxation. arXiv preprint
  arXiv:200904266

\bibitem[{Severn et~al.(2019)Severn, Dryden, and Preston}]{severn2019manifold}
Severn K, Dryden IL, Preston SP (2019) Manifold valued data analysis of samples
  of networks, with applications in corpus linguistics. arXiv preprint
  arXiv:190208290

\bibitem[{Shervashidze et~al.(2011)Shervashidze, Schweitzer, Van~Leeuwen,
  Mehlhorn, and Borgwardt}]{shervashidze2011weisfeiler}
Shervashidze N, Schweitzer P, Van~Leeuwen EJ, Mehlhorn K, Borgwardt KM (2011)
  Weisfeiler-lehman graph kernels. Journal of Machine Learning Research
  12(77):2539--2561

\bibitem[{Shirley and Rushton(2005)}]{shirley2005impacts}
Shirley MD, Rushton SP (2005) The impacts of network topology on disease
  spread. Ecological Complexity 2(3):287--299

\bibitem[{Song et~al.(2013)Song, Fukumizu, and Gretton}]{song2013kernel}
Song L, Fukumizu K, Gretton A (2013) Kernel embeddings of conditional
  distributions: A unified kernel framework for nonparametric inference in
  graphical models. IEEE Signal Processing Magazine 30(4):98--111

\bibitem[{Srivastava and Klassen(2016)}]{FDA}
Srivastava A, Klassen EP (2016) Functional and shape data analysis. Springer

\bibitem[{Sun et~al.(2016)Sun, Kunegis, and Staab}]{konect:sun_2016_49561}
Sun J, Kunegis J, Staab S (2016) Predicting user roles in social networks using
  transfer learning with feature transformation. In: Proc. ICDM Workshop on
  Data Mining in Networks

\bibitem[{Tang et~al.(2015)Tang, Qu, Wang, Zhang, Yan, and Mei}]{tang2015line}
Tang J, Qu M, Wang M, Zhang M, Yan J, Mei Q (2015) Line: Large-scale
  information network embedding. In: Proceedings of the 24th international
  conference on world wide web, International World Wide Web Conferences
  Steering Committee, pp 1067--1077

\bibitem[{Ugander et~al.(2011)Ugander, Karrer, Backstrom, and
  Marlow}]{ugander2011anatomy}
Ugander J, Karrer B, Backstrom L, Marlow C (2011) The anatomy of the facebook
  social graph. arXiv preprint arXiv:11114503

\bibitem[{Umeyama(1988)}]{umeyama1988eigendecomposition}
Umeyama S (1988) An eigendecomposition approach to weighted graph matching
  problems. IEEE transactions on pattern analysis and machine intelligence
  10(5):695--703

\bibitem[{Vayer et~al.(2018)Vayer, Chapel, Flamary, Tavenard, and
  Courty}]{vayer2018optimal}
Vayer T, Chapel L, Flamary R, Tavenard R, Courty N (2018) Optimal transport for
  structured data with application on graphs. arXiv preprint arXiv:180509114

\bibitem[{Vishwanathan et~al.(2010)Vishwanathan, Schraudolph, Kondor, and
  Borgwardt}]{vishwanathan2010graph}
Vishwanathan SVN, Schraudolph NN, Kondor R, Borgwardt KM (2010) Graph kernels.
  Journal of Machine Learning Research 11(Apr):1201--1242

\bibitem[{Vogelstein et~al.(2015)Vogelstein, Conroy, Lyzinski, Podrazik,
  Kratzer, Harley, Fishkind, Vogelstein, and Priebe}]{vogelstein2015fast}
Vogelstein JT, Conroy JM, Lyzinski V, Podrazik LJ, Kratzer SG, Harley ET,
  Fishkind DE, Vogelstein RJ, Priebe CE (2015) Fast approximate quadratic
  programming for graph matching. PLOS one 10(4):e0121002

\bibitem[{Wang et~al.(2016)Wang, Cui, and Zhu}]{wang2016structural}
Wang D, Cui P, Zhu W (2016) Structural deep network embedding. In: Proceedings
  of the 22nd ACM SIGKDD international conference on Knowledge discovery and
  data mining, ACM, pp 1225--1234

\bibitem[{Westenberg et~al.(2008)Westenberg, van Hijum, Lulko, Kuipers, and
  Roerdink}]{westenberg2008interactive}
Westenberg MA, van Hijum SA, Lulko AT, Kuipers OP, Roerdink JB (2008)
  Interactive visualization of gene regulatory networks with associated gene
  expression time series data. In: Visualization in Medicine and Life Sciences,
  Springer, pp 293--311

\bibitem[{White and Wilson(2010)}]{white2010generative}
White D, Wilson RC (2010) Generative models for chemical structures. Journal of
  chemical information and modeling 50(7):1257--1274

\bibitem[{Yanardag and Vishwanathan(2015)}]{yanardag2015deep}
Yanardag P, Vishwanathan S (2015) Deep graph kernels. In: Proceedings of the
  21th ACM SIGKDD International Conference on Knowledge Discovery and Data
  Mining, pp 1365--1374

\bibitem[{Yang(2003)}]{yang2003market}
Yang J (2003) Market segmentation and information asymmetry in chinese stock
  markets: A var analysis. Financial Review 38(4):591--609

\bibitem[{Zhou and De~la Torre(2015)}]{zhou2015factorized}
Zhou F, De~la Torre F (2015) Factorized graph matching. IEEE transactions on
  pattern analysis and machine intelligence 38(9):1774--1789

\end{thebibliography}

\end{document}